\documentclass[runningheads]{llncs}
\usepackage{amsmath,amsfonts}
\usepackage{algorithmic}
\usepackage{algorithm}
\usepackage{array}
\usepackage{subfig}
\usepackage{textcomp}
\usepackage{stfloats}
\usepackage{url}
\usepackage{verbatim}
\usepackage{graphicx}
\usepackage{cite}
\usepackage{caption}
\usepackage{nomencl}
\usepackage{ntheorem}   
\usepackage{lipsum}     
\usepackage{graphicx}
\usepackage{textcomp}
\usepackage{xcolor}
\usepackage{hyperref}
\usepackage{subcaption}

\hypersetup{
	colorlinks=true,
	linkcolor=blue,
	filecolor=green, 
	urlcolor=black,
	citecolor=orange
}

\usepackage{bm}

\DeclareRobustCommand{\uvec}[1]{{%
		\ifcsname uvec#1\endcsname
		\csname uvec#1\endcsname
		\else
		\bm{\mathbf{#1}}%
		\fi
}}
\usepackage{graphicx}
\usepackage{gensymb}

\begin{document}
\title{Development of Underactuated Geometric Compliant (UGC) Module with Variable Radial for Robotic Applications}
\titlerunning{Underactuated Geometric Compliant Module}
%
\author{Mark Krysov\orcidID{0009-0007-4790-3361} \and
Seyed Amir Tafrishi\orcidID{0000-0001-9829-3144}}
\authorrunning{Mark Krysov and Seyed Amir Tafrishi}
%
\institute{School of Engineering, Cardiff University, Cardiff, UK, CF24 3AA  \\
\email{\{krysovm, tafrishisa\}@cardiff.ac.uk}\\
}
\maketitle              
\begin{abstract}
This paper introduces a novel underactuated geometric compliant (UGC) robot and investigates the behaviors of underactuated compliant modules with variable radial stiffness, aiming to enhance the versatility and functionality of UGC robots. We initiate the study by designing and fabricating various compliant semi-rigid geometric joints, each tailored to a specific design objective. These joints undergo physical testing to validate their stiffness characteristics and returnable angles as durability factors. Subsequently, we develop a mathematical model based on Gaussian process regression to incorporate the different geometric joint characteristics, including thickness, facilitating the development of fully functional prototypes with easy-to-3D print models. After analyzing individual joints, we present various configurational combinations to construct the overall UGC module for robotics applications. Our final prototype UGC can dynamically alter its radius, reducing to 80-85\% of its original value while maintaining structural integrity and operational efficiency. This study discusses potential abilities, challenges, and limitations associated with employing UGC modules, offering valuable insights for future research and developments in UGC robotics.
\keywords{Geometric compliant joints  \and Underacuated modular robots \and soft robotics \and Compliant mechanisms design}
\end{abstract}
\section{Introduction}
As the field of robotics increasingly embraces soft and compliant mechanisms, there is a growing recognition of their significance within research studies \cite{manti2016stiffening}. These attributes, characterized by flexibility and compliance, not only enhance the safety of robotic platforms but also afford them greater degrees of freedom in body deployment. However, the attainment of such compliance and flexibility often necessitates intricate and voluminous actuation mechanisms, thereby presenting formidable hurdles for practical implementation in robotic systems. There are challenges in how can robot bodies achieve \textit{modularity} and \textit{compliance} while operating within constraints imposed by a limited number of actuators, known for \textit{underactuation} \cite{liu2020survey}. 

Compliance and softness are crucial for creating safer and more flexible mechanisms \cite{10386077} and robotic systems \cite{rus2015design}, offering advantages across various scenarios. One such scenario is the inspection and maintenance of power plants, where traditional intrusive methods are both difficult and expensive to implement \cite{wang2021robots,10313930}. Utilizing a robot capable of adapting to the environment's shape, like a soft and compliant robot, while also withstanding extreme conditions like a rigid robot, would enable tasks to be performed directly within the system \cite{liu2021review}. This would primarily involve the robot travelling through piping to an area of interest and using its ability to change shape to lock its position so that any needed inspection or repairs can be carried out in a stable environment. With the completion of the task, the robot may then re-form itself and continue to the next task/objective location.  Another potential application is the exploration of cramped and hostile environments, where sharp objects pose a danger to soft robots and challenge traditional wheeled or legged robots' ability to traverse. This would include cave systems for search and rescue or research where using personnel would be unreasonable. However, the challenge lies in both design perspective and material choice \cite{pagoli2021review}, as selecting the appropriate semi-rigid material, such as Polylactic acid (PLA) with easy-to-3D-print models, for compliant robot surface structures remains a significant challenge. Silicon-based materials, although commonly used, are fragile and struggle to endure high force, limiting their suitability for directly controllable surfaces. However, a benefit of the use of compliance is the ability to function at different scales, where drastic redesigns are not required when changing the size of the design, even into microscopic levels.

Another aspect is underactuated robotics that, despite their greater intricacy compared to fully actuated counterparts, offer a multitude of advantages encompassing energy efficiency, material conservation, and space optimization \cite{he2019underactuated,liu2020survey}. The underactuated robots exhibit heightened efficiency and flexibility in select scenarios, particularly when coupled with precise control mechanisms. This efficacy is notably evident in intricate tasks like locomotion \cite{xu2024tracked} or shape transformation \cite{firouzeh2015under} in constrained environments. Underactuated robots play a pivotal role in conservation efforts, as their diminished actuation requirements translate to reduced mass, volume, and energy consumption \cite{he2019underactuated,tafrishi2020singularity,tafrishi2021inverse}. This principle mirrors biological structures in nature; for instance, snakes or birds utilize their multi-skeletal bodies with minimal actuation to execute complex, flexible, and multi-purpose morphological transformations for traversing or moving in space. The primary objective in underactuated systems is to incorporate the least actuator numbers capable of not only adjusting stiffness but also facilitating changes in the body's size or form to navigate challenging environments, akin to the locomotion strategies observed in soft-bodied animals like snakes. However, in the robotics field, this capability is still an open problem of how geometric compliance can be integrated with underactuation to reshape (change size) or transform with a smaller number of actuators.

In this paper, our motivation stems from creating geometrically simple-to-3D print bodies that not only possess compliance and deformability but also utilize underactuated systems to reshape or resize. Therefore, our contributions are as follows:
\begin{itemize}
    \item Conducting an in-depth study on compliant geometric plastic (semi-rigid PLA) joint designs and their ease of 3D printing in Section \ref{Sec:CompliantjointDesign}. Also, analyzing the behavior of joints concerning stiffness and recovery angle.
    \item Developing a model using the Gaussian Regression Process to identify variable stiffness and return angle with varying thickness in Section \ref{Sec:UGCModeling}.
    \item Proposing the final compact printable design, including various module design variations, of an underactuated geometric compliant (UGC) module with a motor actuator in Section \ref{Sec:UGCCompliantmodule}.
\end{itemize}
\section{Geometric Compliant Joint Design and Tests}
\label{Sec:CompliantjointDesign}
This section delves into fundamental studies on compliance and flexibility across various geometric joint designs. By examining how geometry impacts joint stiffness and recovery, we lay the groundwork for understanding the optimal combinations necessary to achieve a variable radius module design. Also, we explore how it can be easy to 3D print the whole module. 
\begin{figure}[t!]
    \centering
     \subfloat[\centering Straight]{\includegraphics[width=0.2\linewidth]{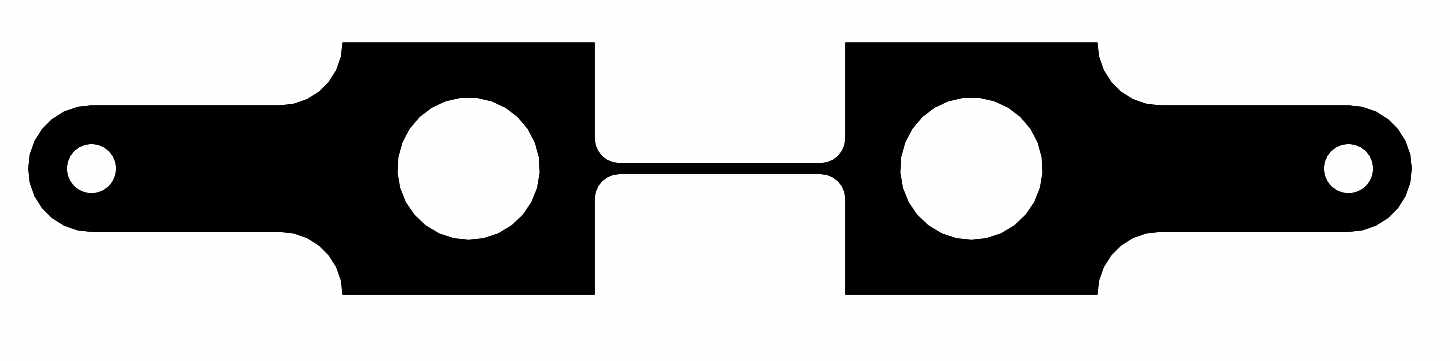}}%
    \qquad
    \subfloat[\centering 0.4mm curve design]{{
  \includegraphics[width=0.2\linewidth]{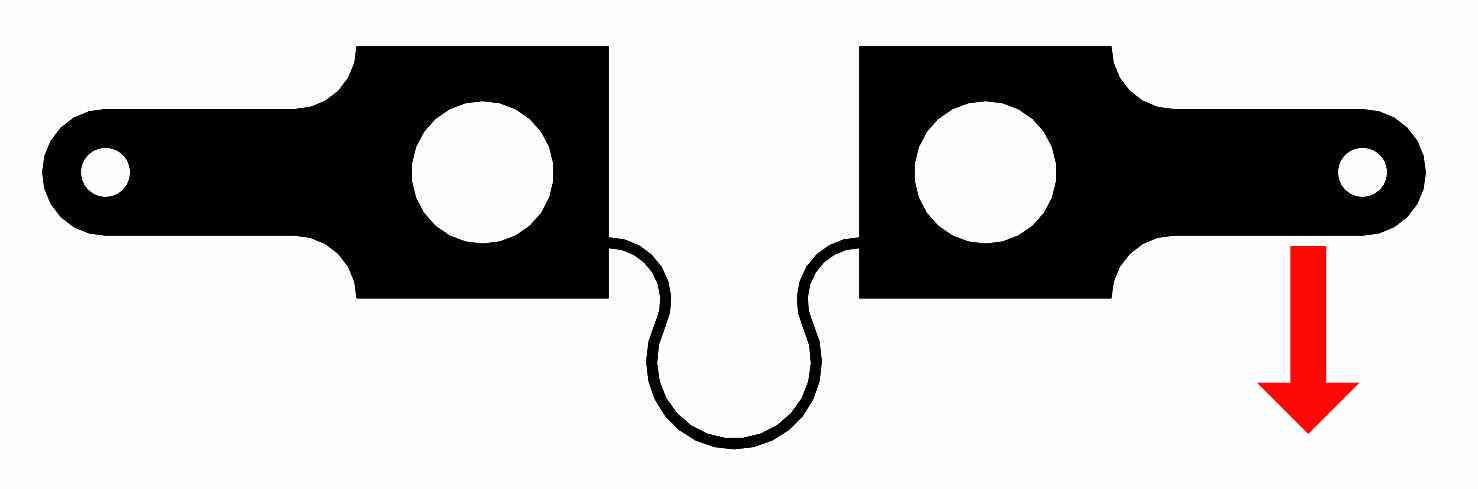}}}
    \qquad
    \subfloat[\centering 0.8mm curve design]{{    \includegraphics[width=0.2\linewidth]{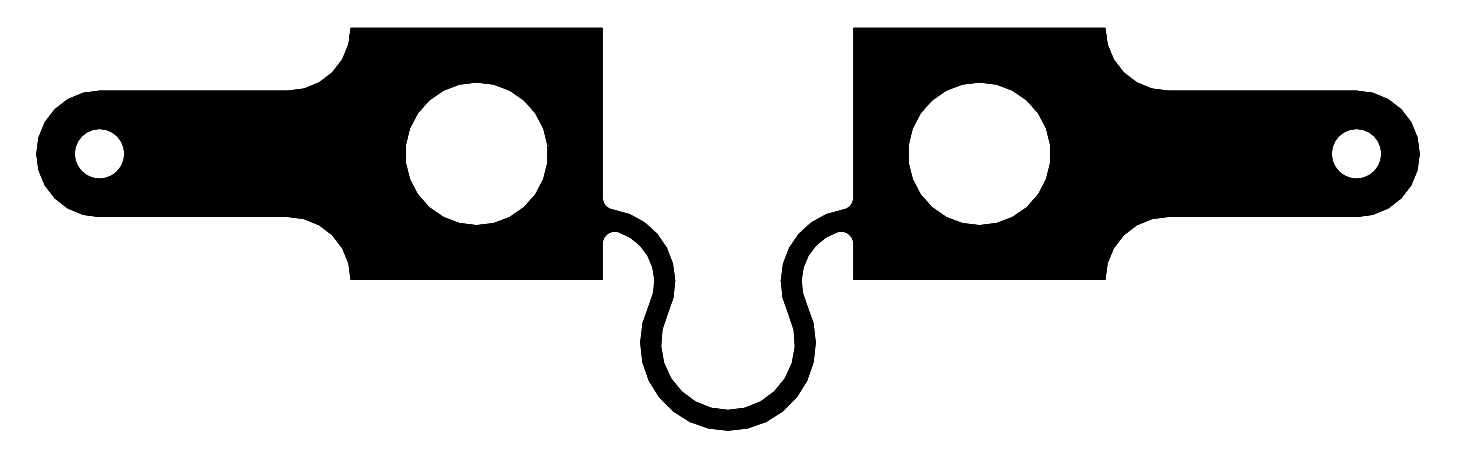}}}
    \qquad
     \subfloat[\centering 1.6mm curve design]{{         \includegraphics[width=0.2\linewidth]{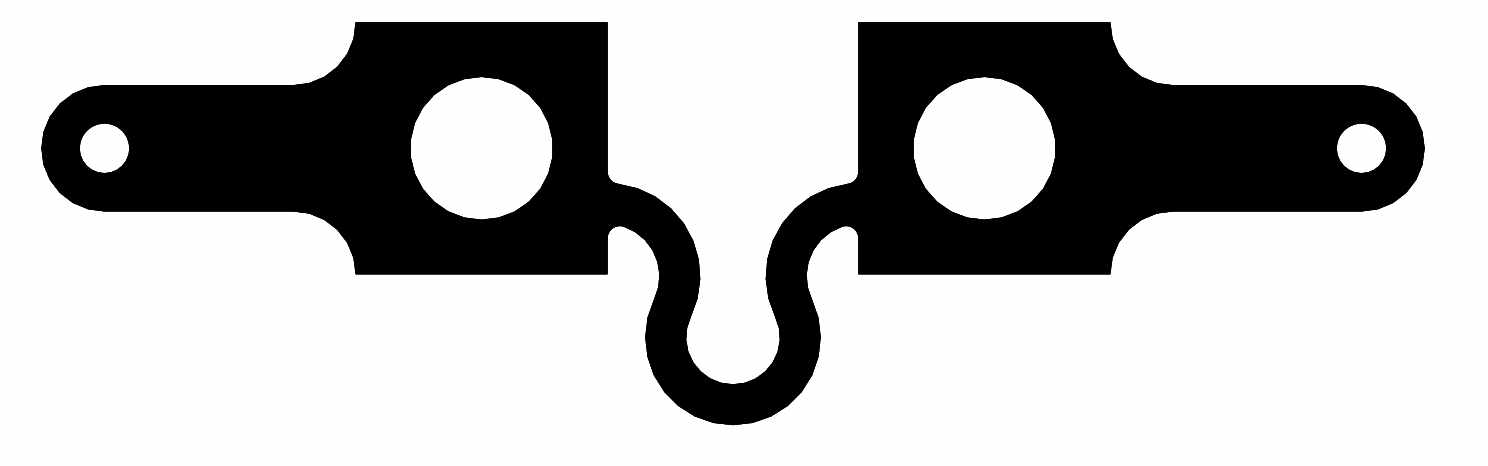}}}
     \qquad
     \subfloat[\centering Double joint design]{{           \includegraphics[width=0.2\linewidth]{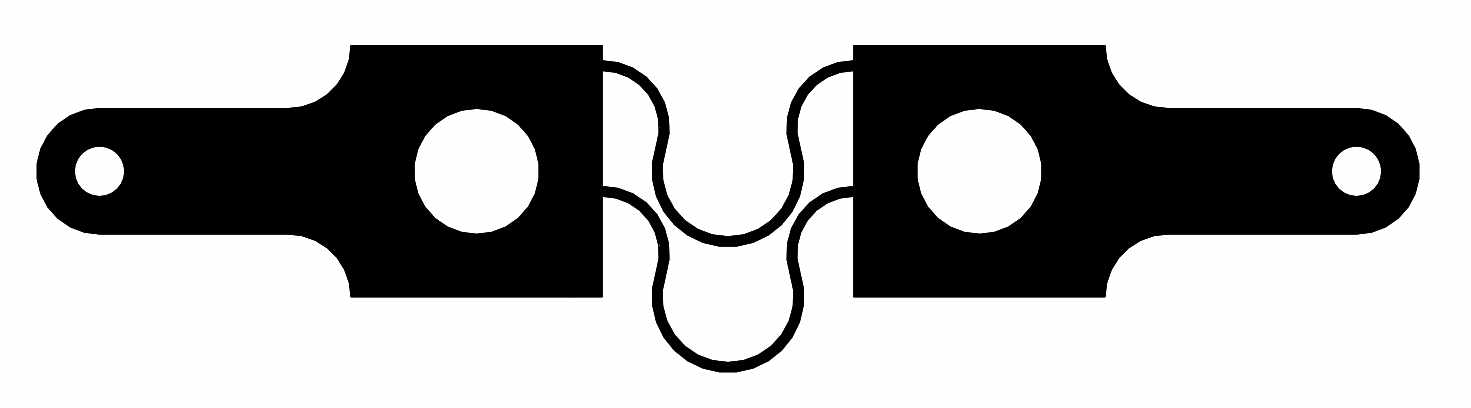}}}
     \qquad
     \subfloat[\centering Symmetrical square Wave]{{           \includegraphics[width=0.2\linewidth]{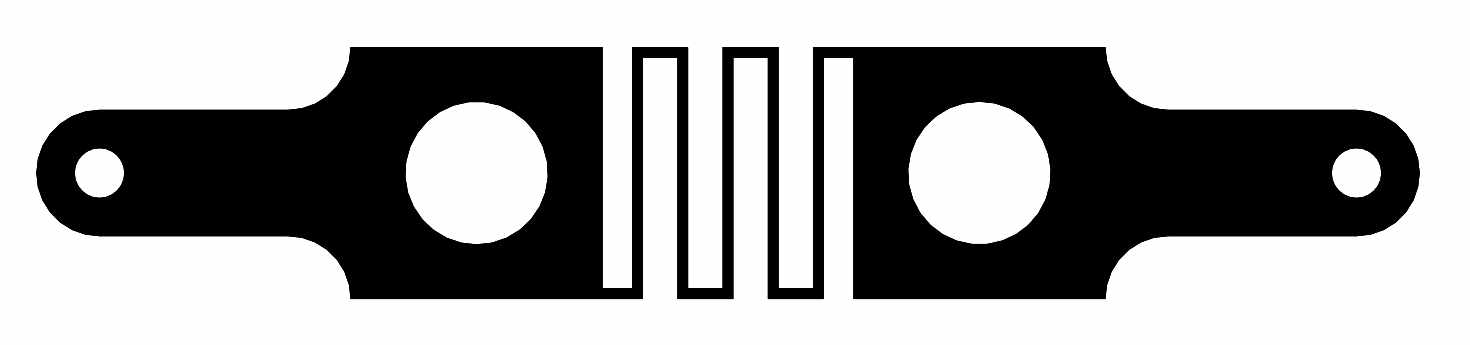}}}
     \qquad
     \subfloat[\centering Non-symmetrical square Wave]{{           \includegraphics[width=0.2\linewidth]{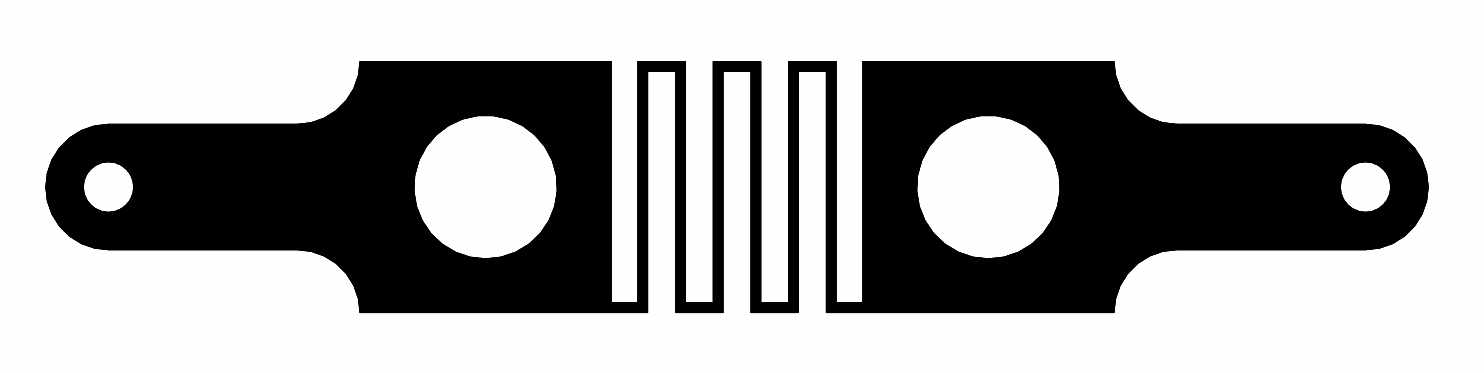}}}
    \caption{Series of designed geometric compliant joints.}
    \label{fig:allthumbnails} 
\end{figure}
 \begin{figure}[t!]
    \centering
     \subfloat[\centering Measurement test bed]{{           \includegraphics[width=0.52\linewidth]{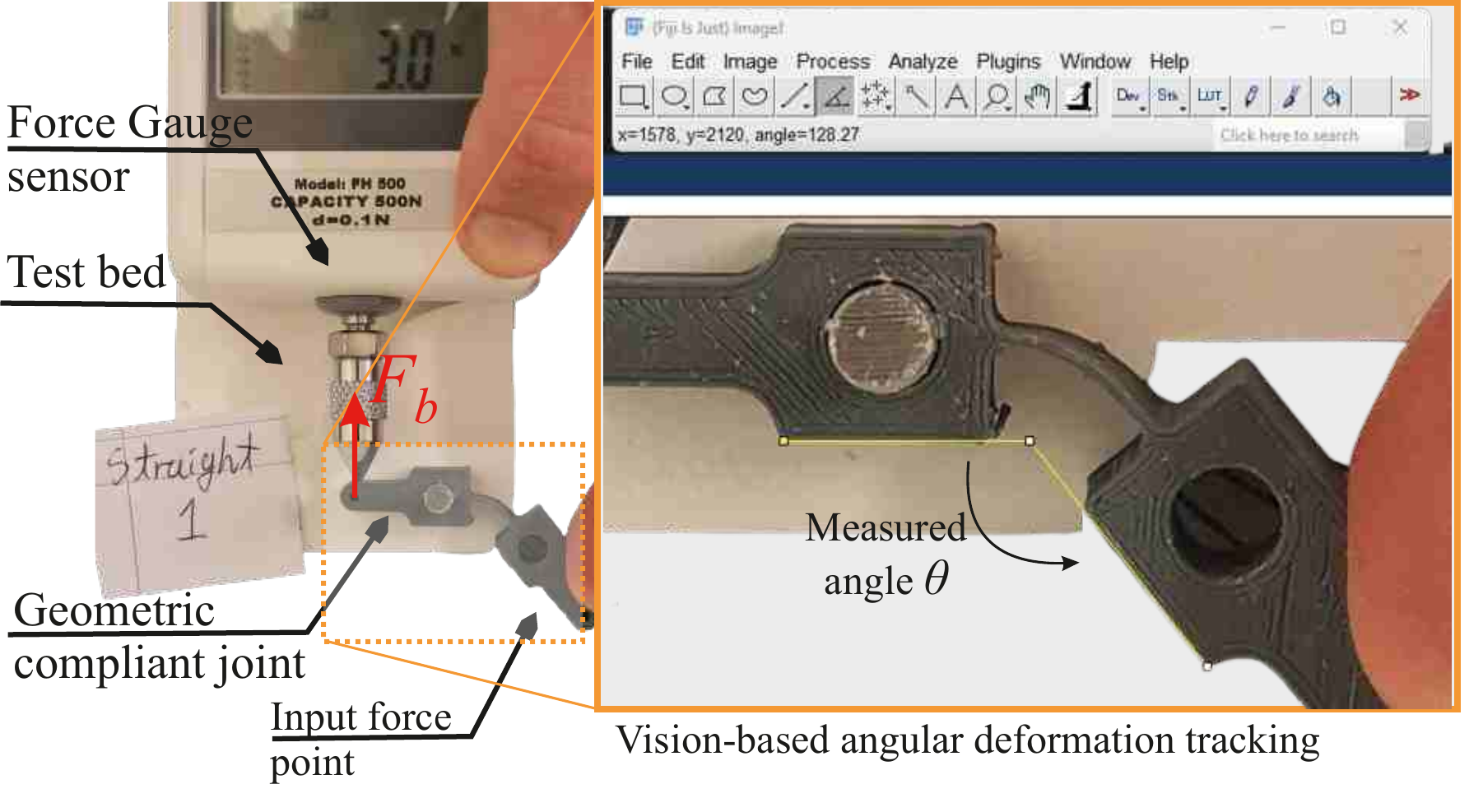}}}\qquad
     \subfloat[\centering Multi-directional bending]{{    \includegraphics[height=3.2cm]{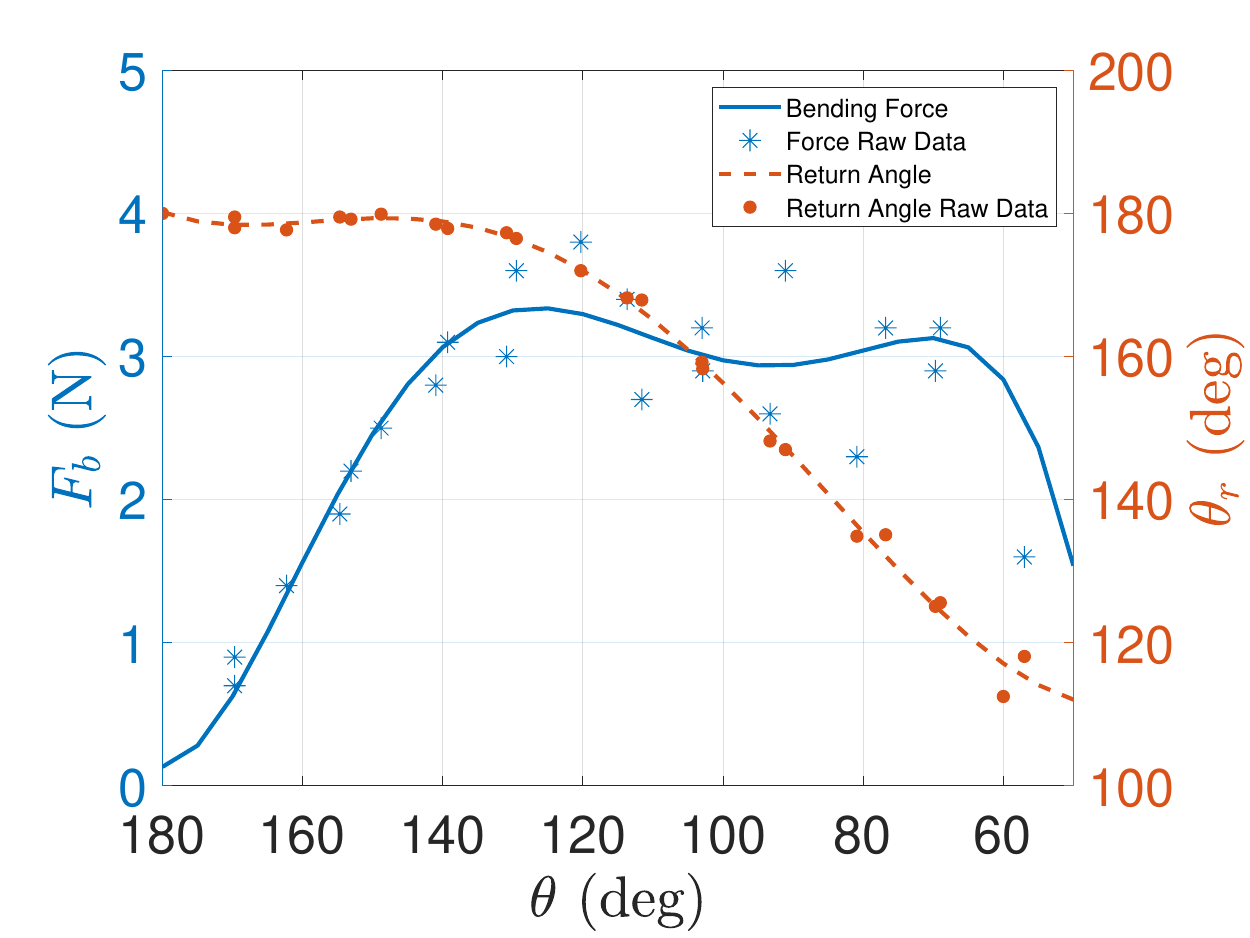}}}
    \caption{(a) The measurement of each joint for testing stiffness and return angle,(b) The force (blue line) and return angle (red line) versus the deformation angle for the straight joint.}
    \label{fig:straightbarjoint}
\end{figure}

We have considered a variety of geometric joint forms, as shown in Fig. \ref{fig:allthumbnails}, fabricated using 3D printed PLA materials. Each joint is measured and analyzed based on the configuration depicted in Fig. \ref{fig:straightbarjoint}-(a). During this process, the force exerted on each designed geometric compliant joint is measured using a Force gauss sensor placed at one end of the joint. Additionally, we measure the angular bending $\theta$ using computer vision techniques, specifically ImageJ Fiji. The depicted first straight linear connection, shown as a joint in Fig. \ref{fig:allthumbnails}-(a), displayed a material stress-strain relationship \cite{plasticstressstrain}, resulting in a predictable yield point. Fig. \ref{fig:straightbarjoint}-(b) illustrates the stiffness change concerning force $F_b$ and return angle $\theta$, along with the recovery/return angle $\theta_r$ after the impacted force. The measured data are averaged and fitted with high-order polynomial curves to illustrate the overall behaviour of stiffness and return angle. Beyond an angle of $\theta=135$\degree, the joint exhibited plastic deformation, impeding its return to the full recovery angle of  $\theta_r=$180\degree. This premature onset of plastic deformation restricts joint motion and heightens the risk of fatigue damage, potentially leading to premature failure. Furthermore, the test revealed uneven deformation and the formation of a crease off-center, shifting the joint's rotation center during testing. While the initial stages of the test demonstrated consistent force requirements to reach specific angles, post-yield point data displayed significant variability, signifying unpredictability and rendering it unsuitable for the design phase.
 
\begin{figure}[t!]%
    \centering
    \subfloat[\centering Aligning the red arrow]{{\includegraphics[height=3cm]{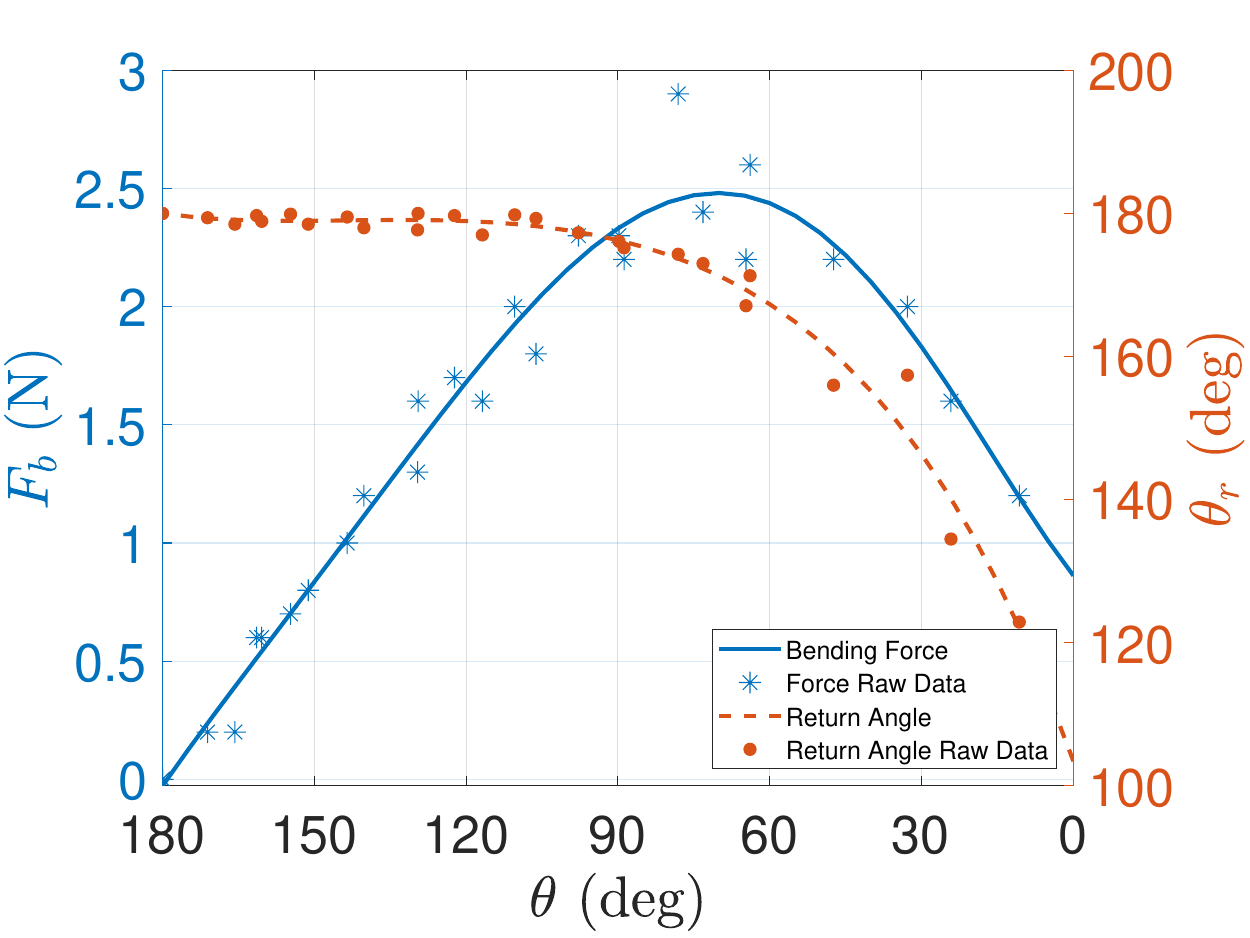} }}%
    \qquad
    \subfloat[\centering Opposite direction]{{\includegraphics[height=3cm]{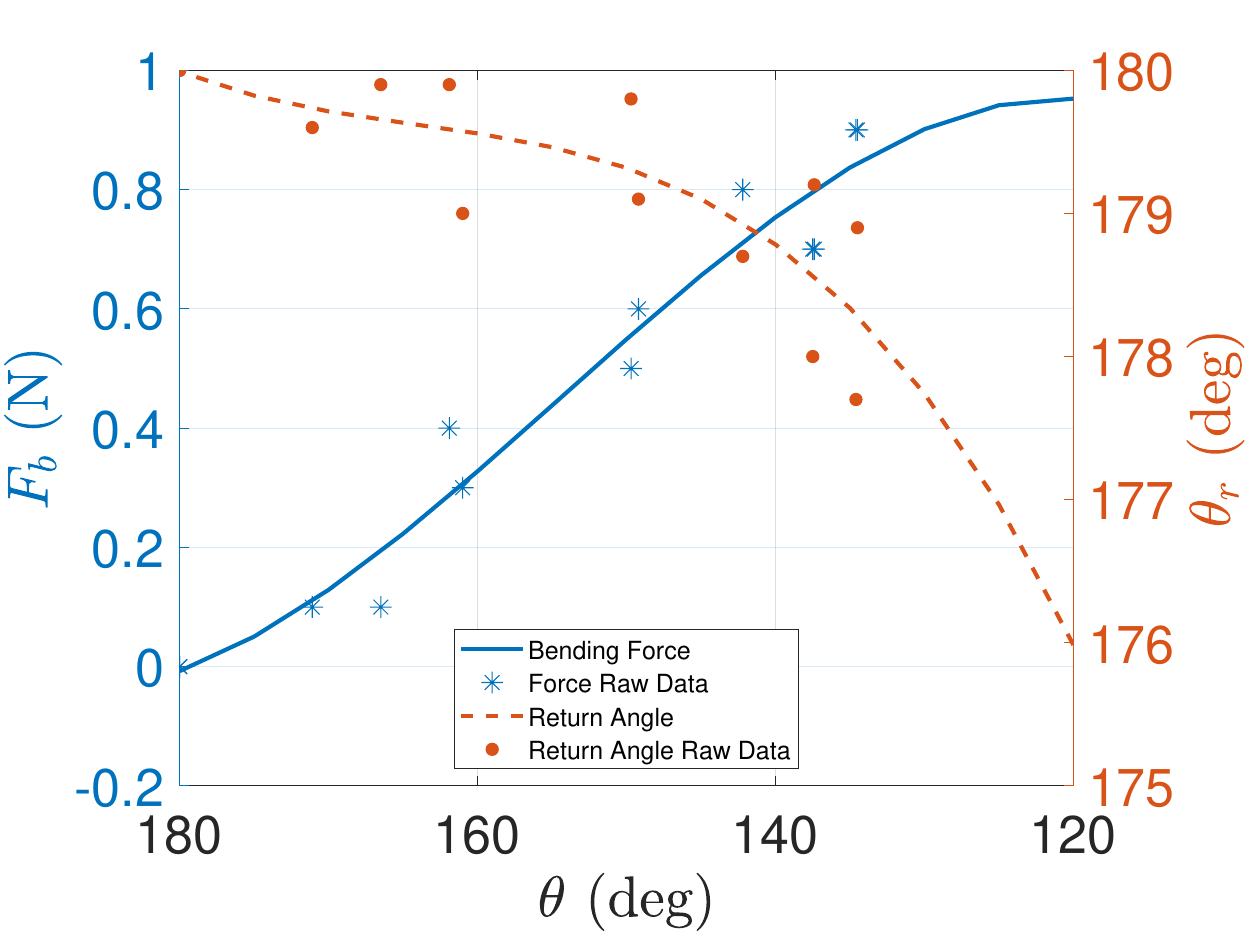} }}%
    \subfloat[\centering ]{{\includegraphics[height=0.9cm, angle=90]{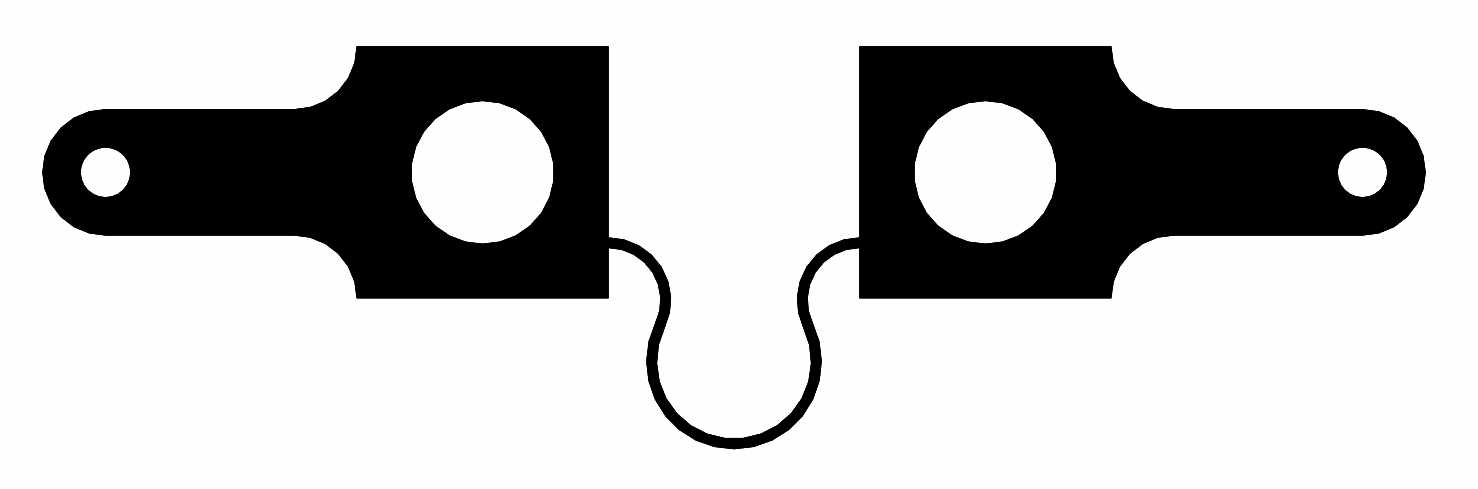} }}%
    \caption{The force (blue line) and return angle (red line) versus the deformation angle for $T=$ 0.4 mm curve joint.}%
    \label{fig:wave4mmprototype}%
\end{figure}
\begin{figure}[t!]%
    \centering
    \subfloat[\centering Aligning the red arrow]{{\includegraphics[height=3cm]{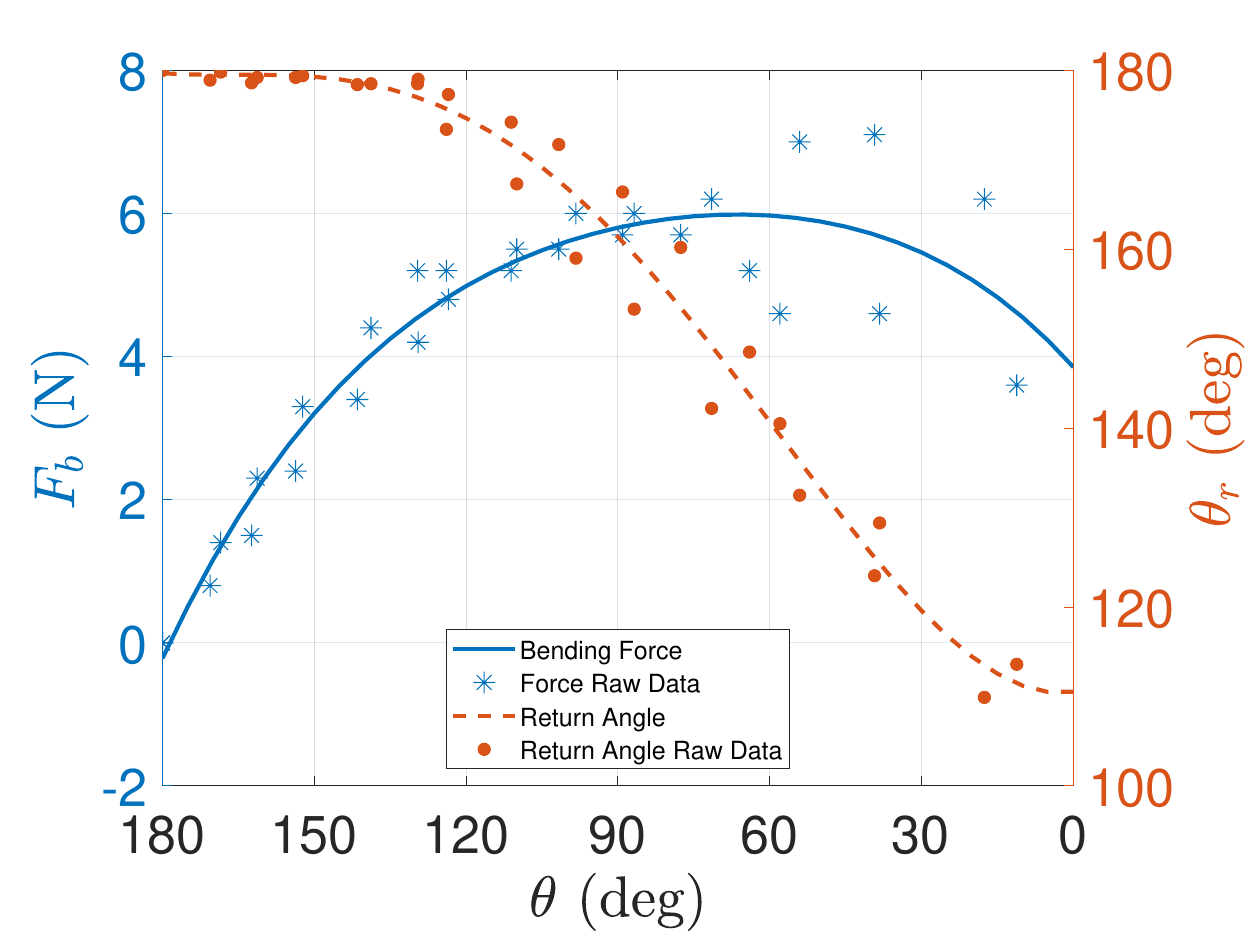} }}%
    \qquad
    \subfloat[\centering Opposite direction]{{\includegraphics[height=3cm]{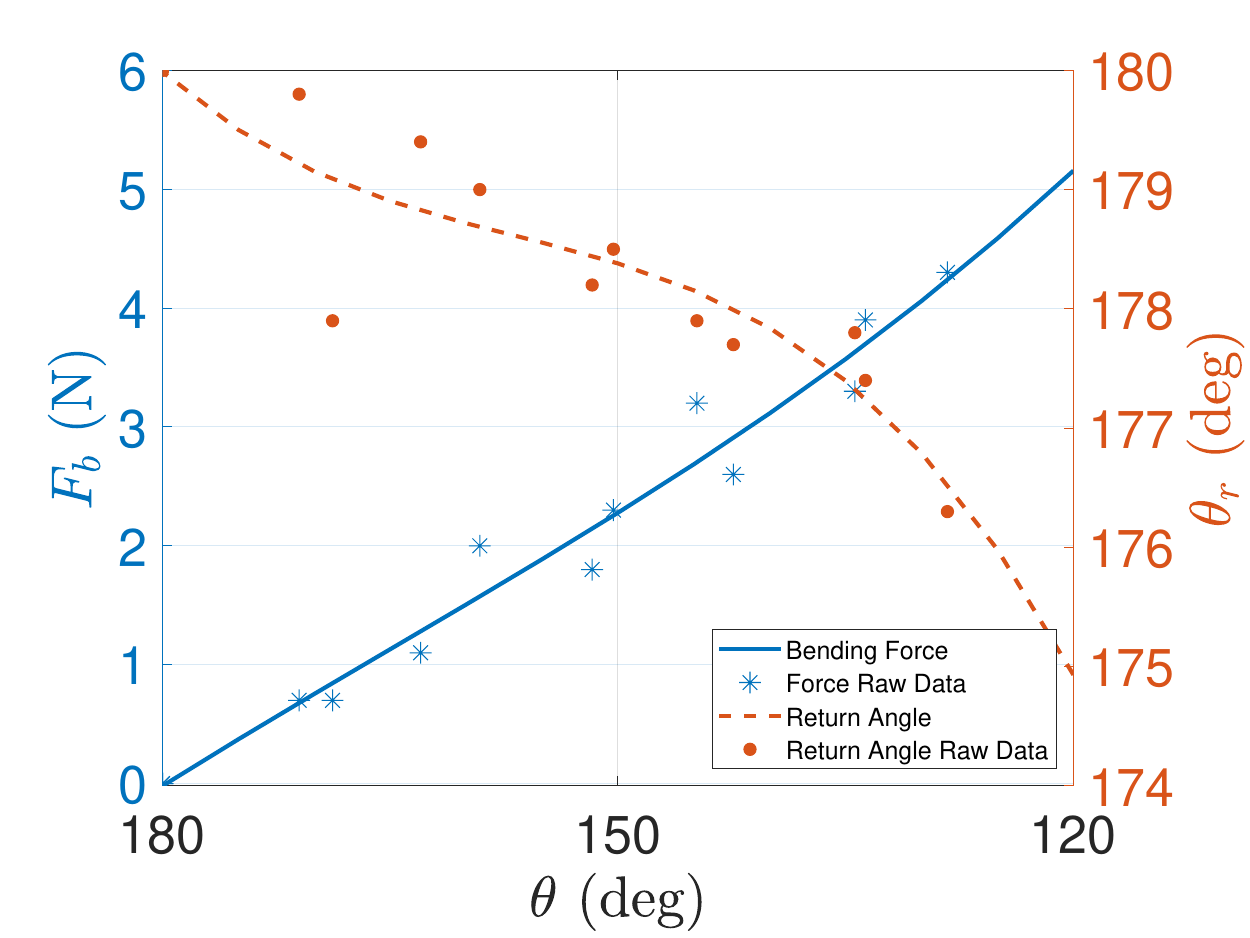} }}%
    \subfloat[\centering ]{{\includegraphics[height=0.85cm, angle=90]{0.8mmcurvethumbnail.jpg} }}%
    \caption{The force (blue line) and return angle (red line) versus the deformation angle for $T=$ 0.8 mm curve joint.}%
    \label{fig:wave8mmprototype}%
\end{figure}
The second geometric joint, the decentralized curved (wave) with a $T=0.4$ mm thickness design (Fig. \ref{fig:allthumbnails}-(b)), employs a strategy of increasing material separation between plates in one direction while eliminating sharp corners to reduce stress concentrations and potential failure. This approach effectively extends the straight joint, distributing force over a larger material area, thereby reducing stress on individual joint elements. Fig. \ref{fig:wave4mmprototype} illustrates the behaviour of this joint in both directions, demonstrating prolonged angle preservation up to approximately $\theta=90$\degree and a linear stiffness pattern, enabling a wide range of motion. Additionally, it requires low force, with a maximum of $F_b=2.9$ N. Reversing the bending direction yields a similar response, with tests concluding due to template self-contact. Consistency between test runs, even beyond the yield point, confirms stable repeatability.

 \begin{figure}[t!]%
    \centering
    \subfloat[\centering Aligning the red arrow]{{\includegraphics[height=3cm]{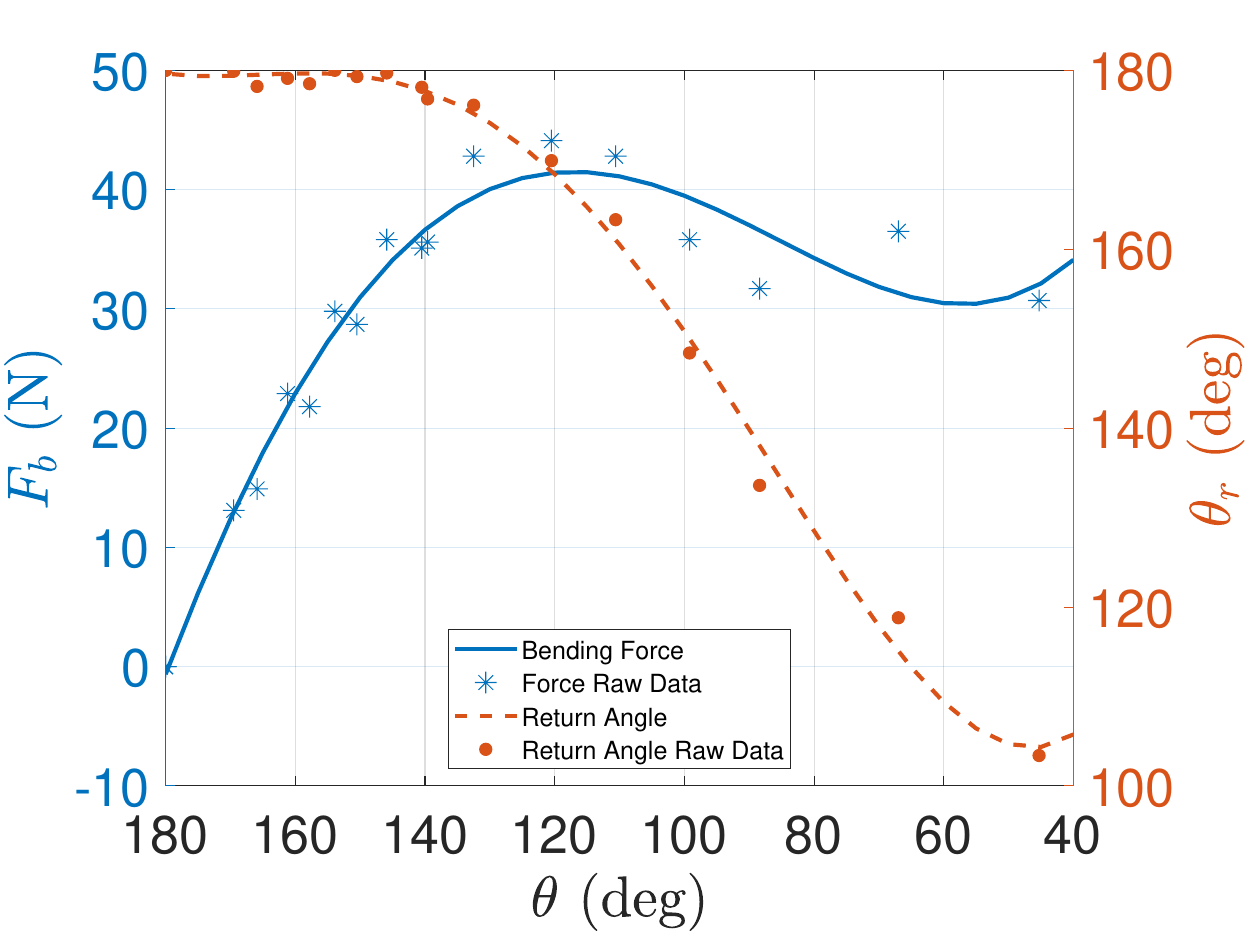} }}%
    \qquad
\subfloat[\centering Opposite direction]{{    \includegraphics[height=3cm]{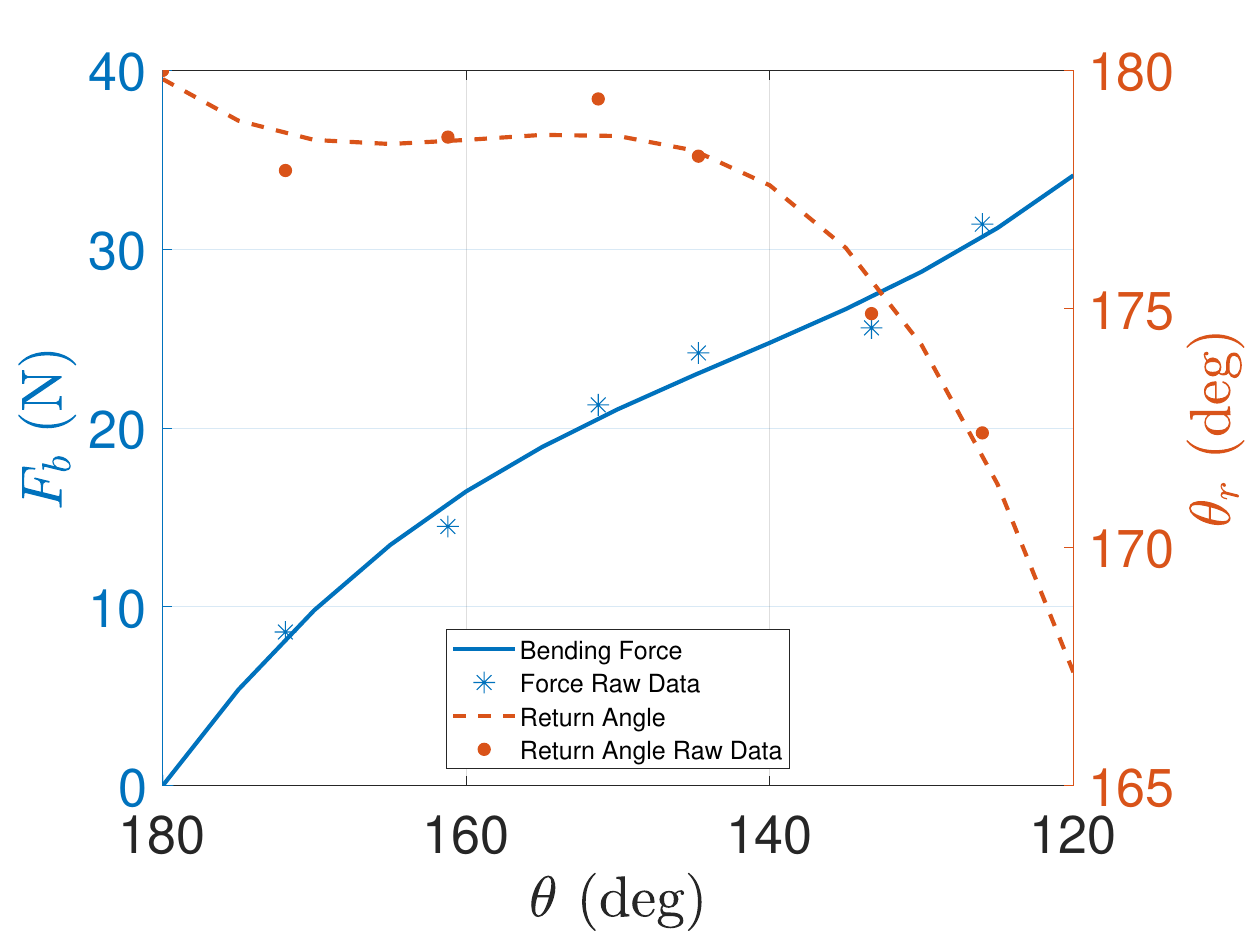} }}%
\subfloat[\centering ]{{\includegraphics[height=0.9cm, angle=90]{1.6mmcurvethumbnail.jpg} }}%
    \caption{The force (blue line) and return angle (red line) versus the deformation angle for $T=$ 1.6 mm curve joint.}%
    \label{fig:wave16mmprototype}%
\end{figure}

The next two joints (Fig. \ref{fig:allthumbnails}(c)-(d)) belong to the same family as $T=$ 0.4 mm curve joint, with increasing thicknesses. Testing aimed to understand how joint behaviour changes with thickness variation, allowing for fine-tuning of compliant modular robots' responses via localized resistances. Thicker joint designs (Figs. \ref{fig:wave8mmprototype}-\ref{fig:wave16mmprototype}) exhibited higher force requirements, reaching a maximum of $F_b=$7.1 N, and yielded at around $\theta$ = 140\degree, earlier than expected. The response was similar in the reverse direction, but beyond the yield point, results showed deviation, indicating some inconsistency between tests. Delamination of the joint connection, consisting of two wall layers due to a 0.4mm print nozzle, could be a contributing factor. The return angle starts to present smaller changes with linear form as the layers get thicker in joint design.

 \begin{figure}[t!]%
    \centering
    \subfloat[\centering Aligning the red arrow]{{\includegraphics[height=3cm]{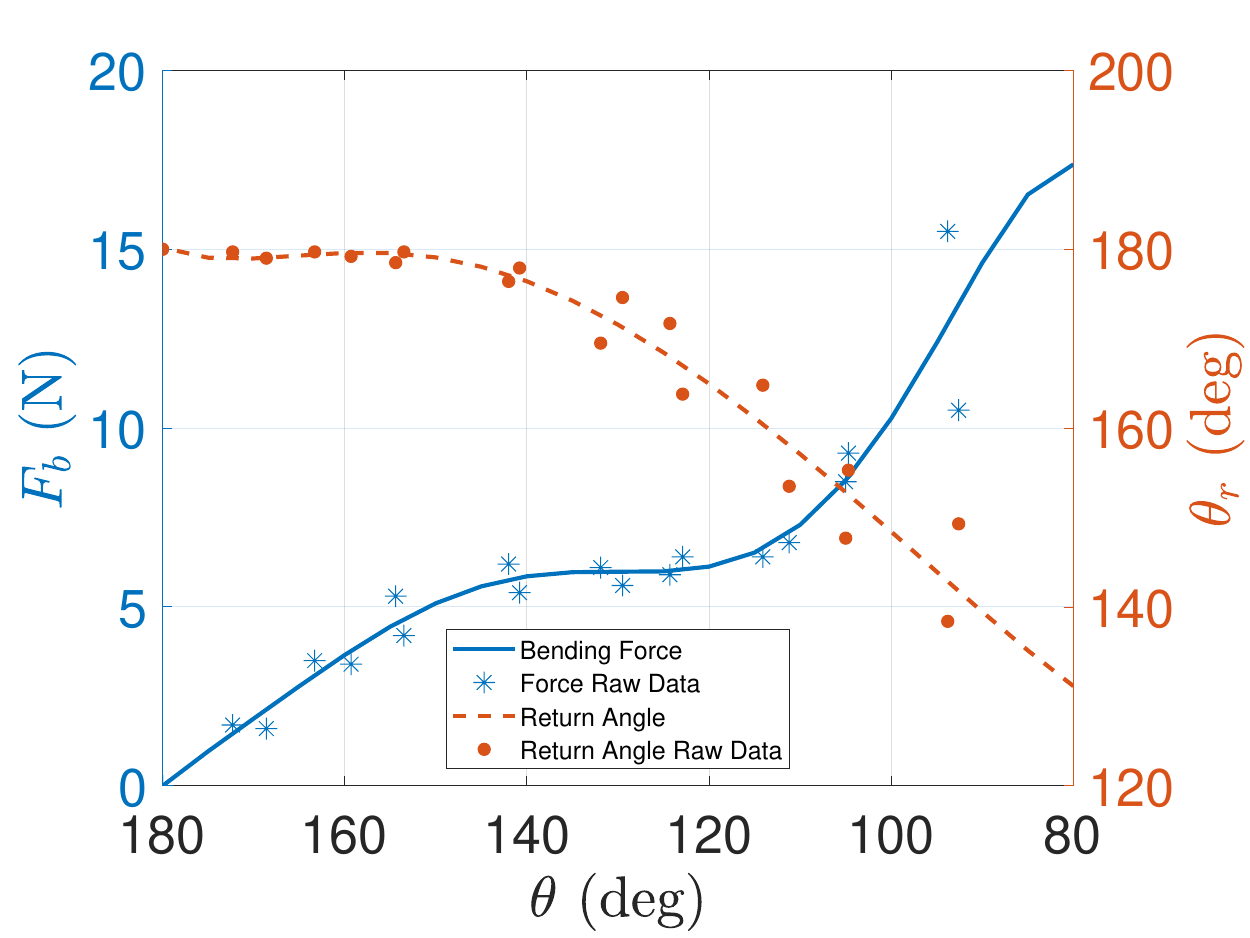} }}%
    \qquad
\subfloat[\centering Opposite direction]{{    \includegraphics[height=3cm]{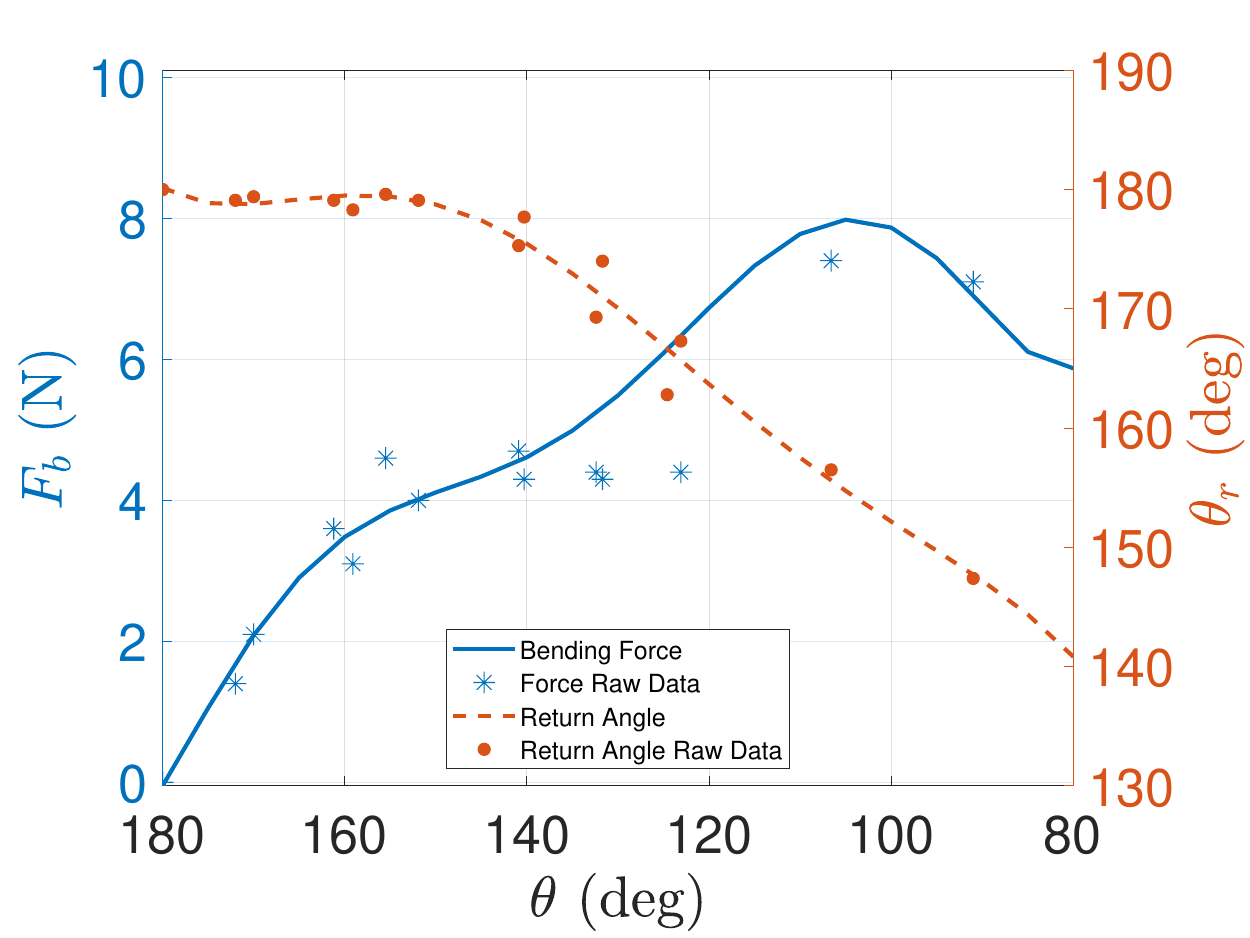} }}%
\subfloat[\centering ]{{\includegraphics[height=0.75cm, angle=90]{doublecirvethumbnail.jpg} }}%
    \caption{The force (blue line) and return angle (red line) versus the deformation angle for the double curve joint.}%
    \label{fig:doublecurvedesing}%
\end{figure}
 \begin{figure}[t!]%
    \centering
    \subfloat[\centering Multi-directional]{{\includegraphics[height=2.4cm]{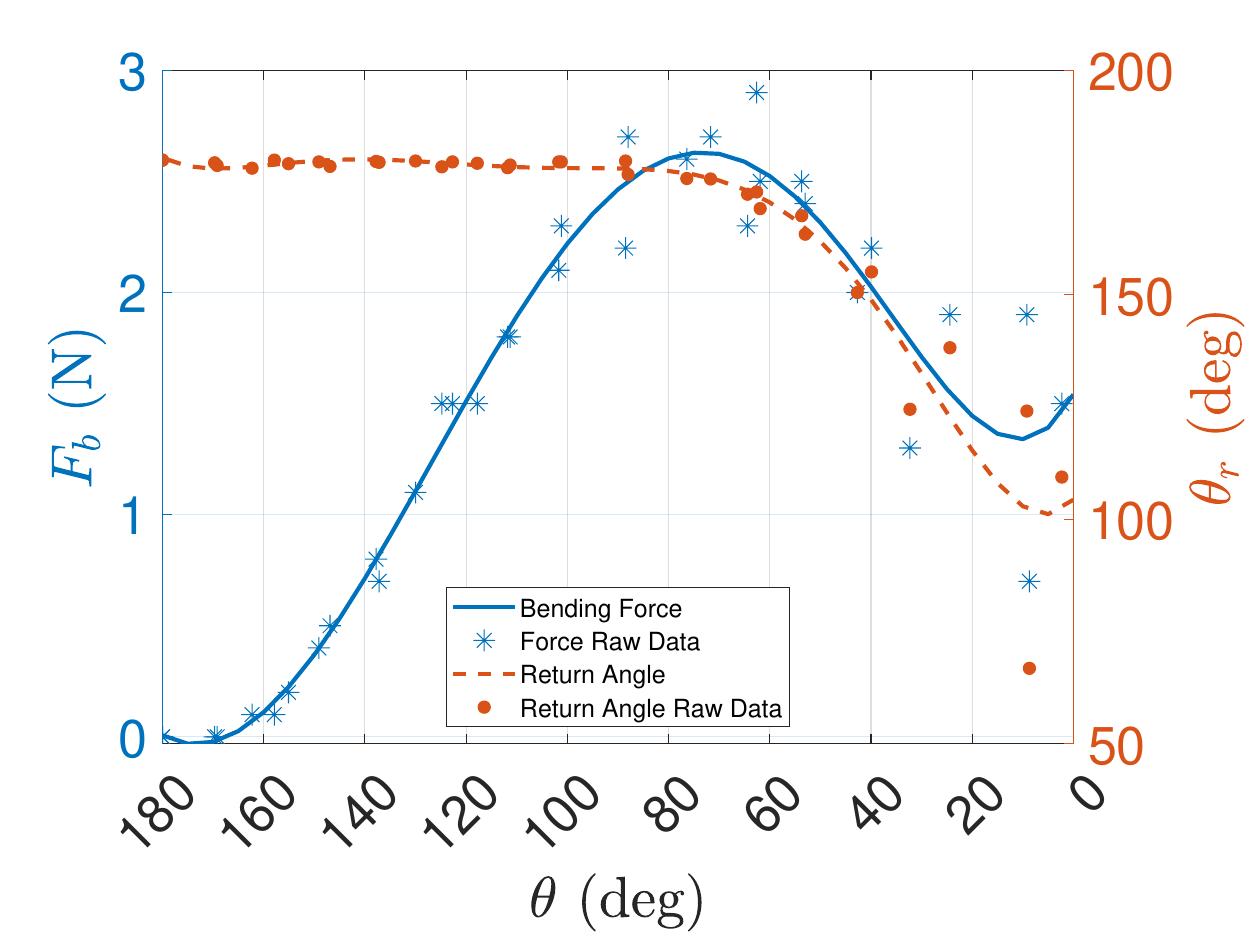} }}%
    \qquad
\subfloat[\centering Aligning the arrow]{{    \includegraphics[height=2.4cm]{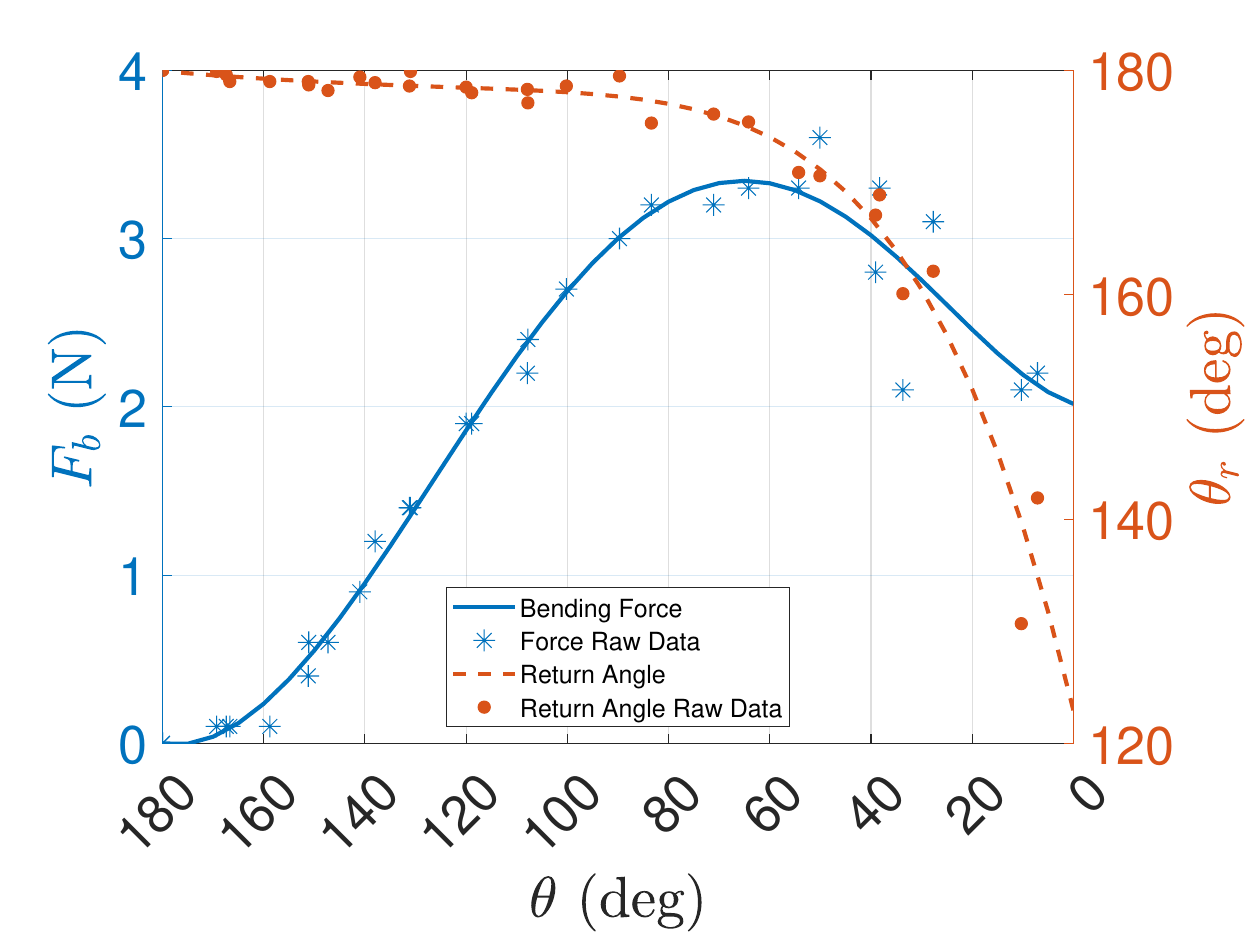} }}%
    \qquad
\subfloat[\centering Opposite direction]{{    \includegraphics[height=2.3cm]{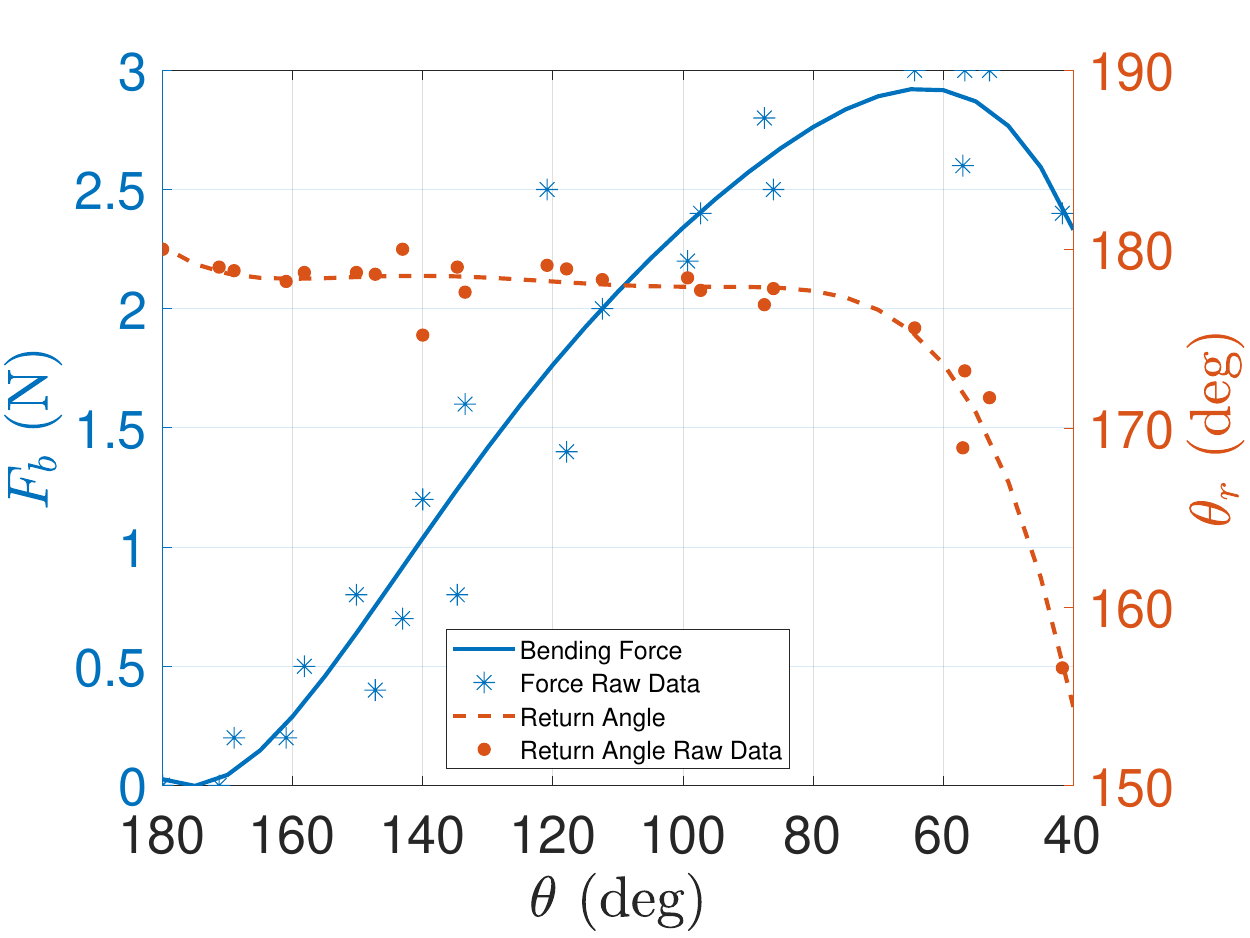} }}
\subfloat[\centering ]{{\includegraphics[trim=20 30 20 30,clip,height=0.4cm, angle=90]{wigglethumbnail.jpg} }}%
    \caption{(a) The force (blue line) and return angle (red line) versus the deformation angle for the symmetrical square wave joint, (b)-(c) The force (blue line) and return angle (red line) versus the deformation angle for the non-symmetrical wave.}%
    \label{fig:squarewavedeisngjointres}%
\end{figure}

The double curve joint with a parallel design (Fig. \ref{fig:allthumbnails}(e)) reached its yield point at around $\theta=$150\degree as shown in Fig. \ref{fig:doublecurvedesing}. Its peak force is challenging to quantify as it typically operates at $F_b=$ 6.8 N. However, due to self-contacting at $\theta$ = 110\degree, the force dramatically increases to $F_b=$ 15.5 N. This joint also experienced early damage, with the central joint being pulled apart unevenly. In one test, the proximity of the two curves caused them to fuse during printing, effectively creating a straight joint, and leading to additional damage.

Inspired by studies on the behaviour of square-based waveforms \cite{howell2013handbook} (Fig. \ref{fig:allthumbnails}(f)-(g)), we integrated this new joint type into our analysis. Here, "non-symmetrical joint" refers to directional bending not occurring at the same connection point upon switching. This joint style exhibited exceptional performance, showing a gradual return angle decrease only starting at $\theta=$70\degree (Fig. \ref{fig:squarewavedeisngjointres}(a)), allowing it to reach challenging angles without fatigue. It requires low actuation force, facilitating motor operation, and demonstrates near-linearity (prior to the yield point), simplifying prediction. Consistency across tests and behaviour suggests its reliability, potentially making it the most optimal joint design. The joint functions normally without interference despite self-contact starting at $\theta$ = 150\degree as shown in Fig. \ref{fig:squarewavedeisngjointres}(b)-(c). Compared to symmetrical and non-symmetrical joints, the nonlinearity of the latter presents connections from different ends, with a larger return angle at around $\theta=40$\degree.
\section{UGC Mechanism Modeling}
\label{Sec:UGCModeling}
\subsection{Stiffness and Return Angle Modeling of Joints}
In this section, we model the stiffness and return angle of the targeted joints using the Gaussian Process Regression (GPR) method. This GPR allows us to have two models the multi-input nonlinear model: a) inputs as the thickness and angle and output as the force. b) input was the thickness and given angle and output was potential recovery/return angle. Note that polynomial curve fitting limitations necessitate the shift to use Gaussian regression since it offers greater accuracy in value prediction. This enables precise calculation of force requirements across various designs, ensuring optimal functionality. 

Consider a training set $\{(x_i,y_i); i=1,2,...,n\}$, where $x_i \in \mathbb{R}^{d\times d}$ and $y_i \in \mathbb{R}$, drawn from an unknown distribution for the training data \cite{rasmussen2006gaussian}. The GPR model of the form is considered: first case as $x=\left[T,\;\theta\right],\;y_1=F_b$, the second case with an output of $y_2=\theta_r$ in the following form
\begin{equation}
    y_i = x^T \beta + \varepsilon,
    \label{Eq:Gaussian_reg}
\end{equation} 
where $\varepsilon \sim \mathcal{N}(0,\sigma^2)$. The error variance $\sigma^2$ and the coefficients $\beta$ are estimated from the data. The GPR has a set of random variables such that any finite number of them have a joint Gaussian distribution. GP function is defined by its mean function $m(x)$ and covariance function $k(x,x')$ which is the Kernel-based covariance matrix in our study and is computed as squared exponentials. For the GPR model in (\ref{Eq:Gaussian_reg}), from Matlab toolbox gave us the $\beta=[-2.4933, 
    0.1164,
         0,
   -0.0007,
    8.4377]$ and $\varepsilon=1.9272
$ model parameters with covariance matrix below $10^{-3}$ values for the wave joints. And, we obtained the following $\beta=[
    1.6940,\;
    0.0225,\;
   -0.0002]$ and $\varepsilon=0.2916$ values for the square joints. Note that the Gaussian regression model for the curve design is more complicated as it has an extra dimension of thickness that was measured from $T\in[0.4,1.6]$ mm in 0.4 mm steps.


\begin{figure}[t!]%
    \centering
    \subfloat{{\includegraphics[height=3.1cm]{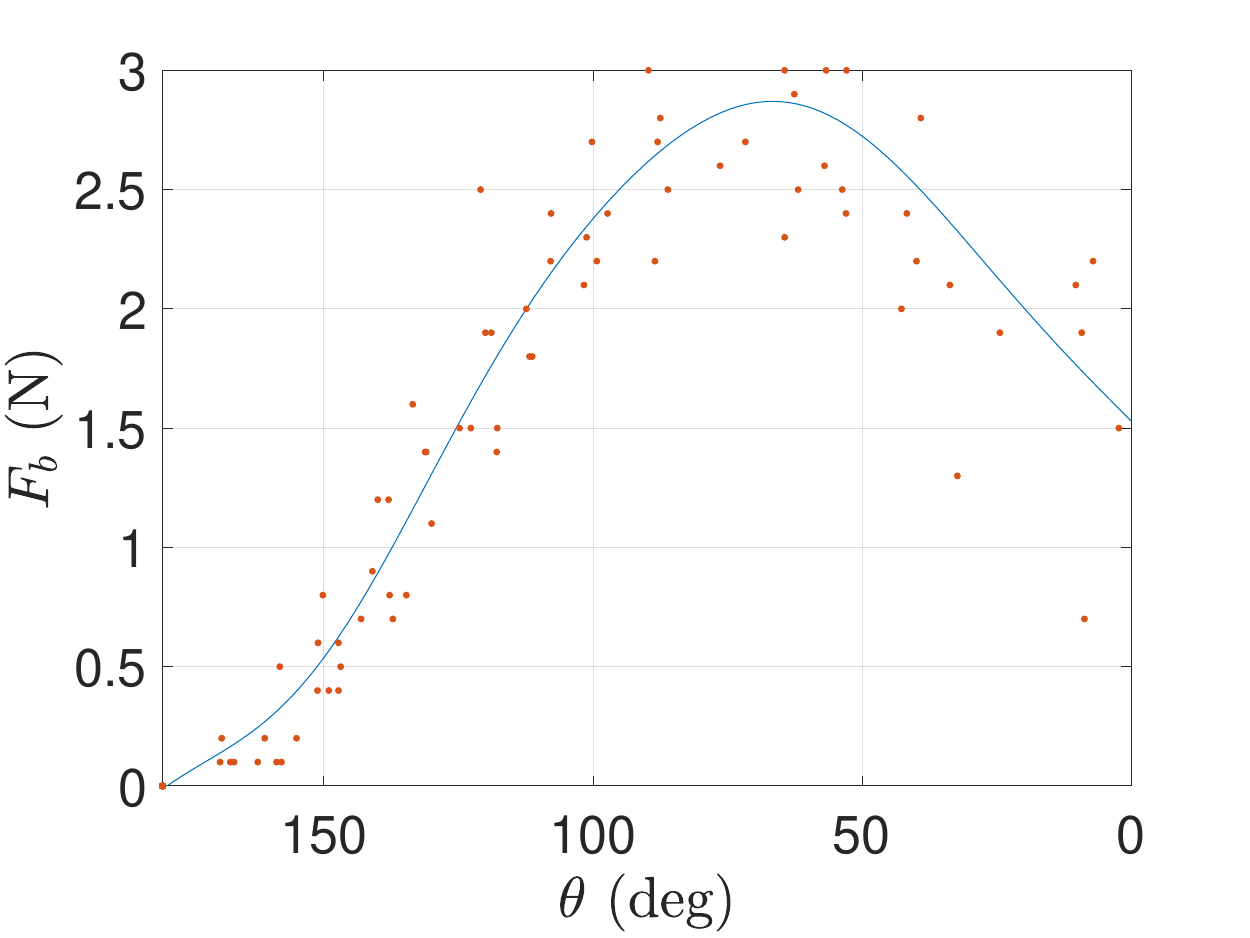} }}%
    \qquad
    \subfloat{{\includegraphics[height=3.1cm]{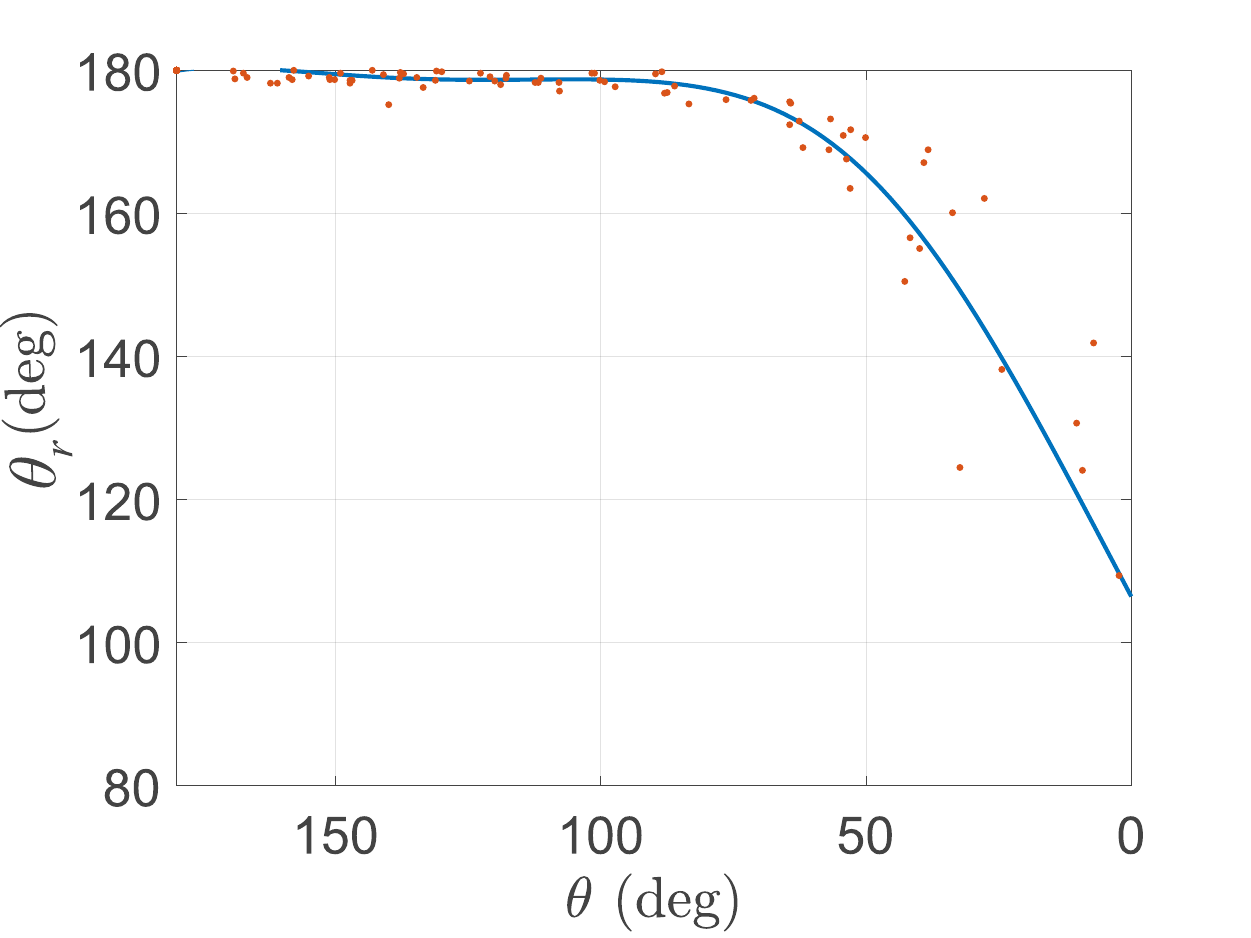} }}%
    \caption{The trained GP model for angle input $\theta$ and output of body force $F_b$ and return angle $\theta_r$ of square wave joint.}%
    \label{fig:gaussiamwigglereturn1}%
\end{figure}
\begin{figure}[t!]%
    \centering
    \subfloat{{\includegraphics[height=3cm]{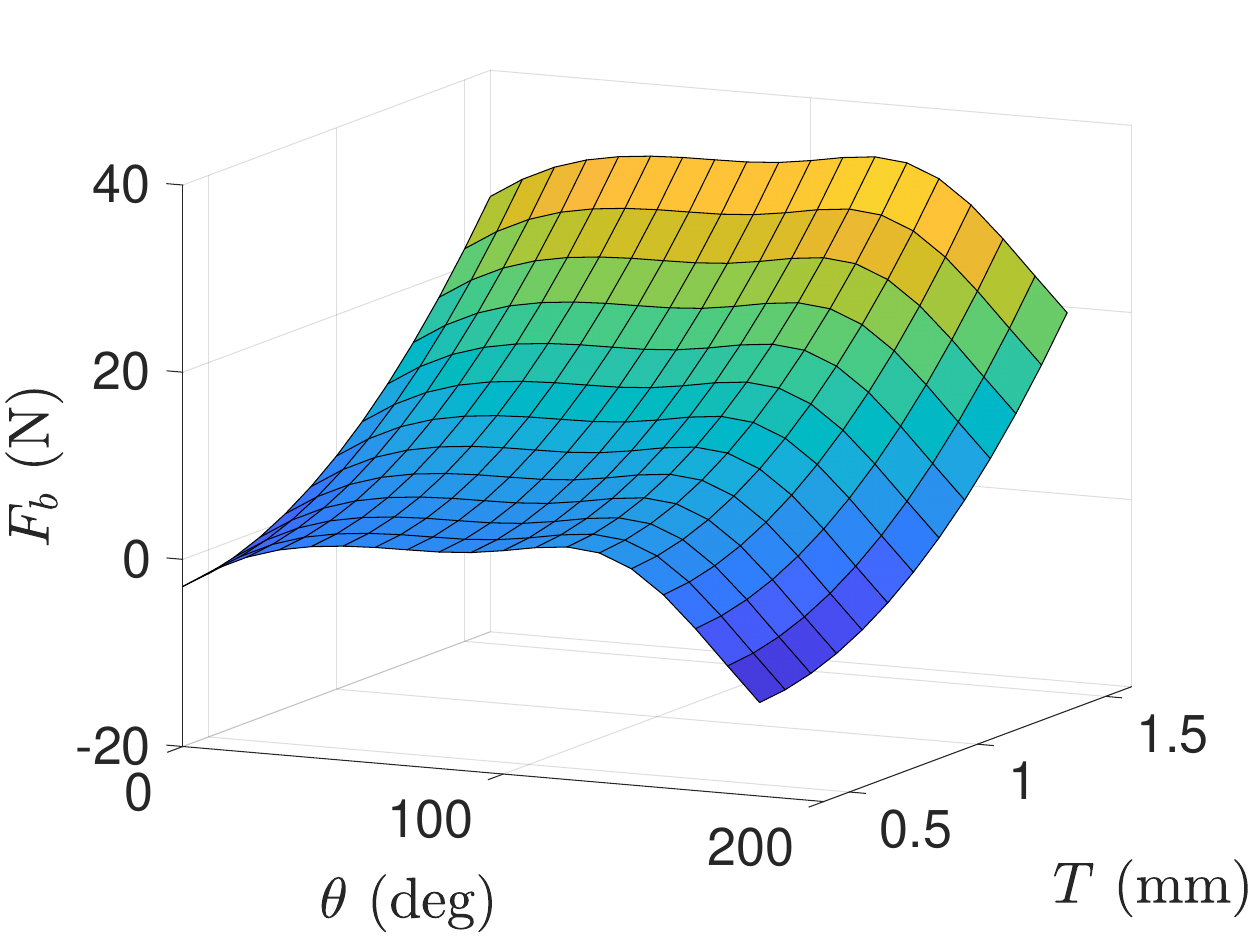} }}%
    \qquad
    \subfloat{{\includegraphics[height=3cm]{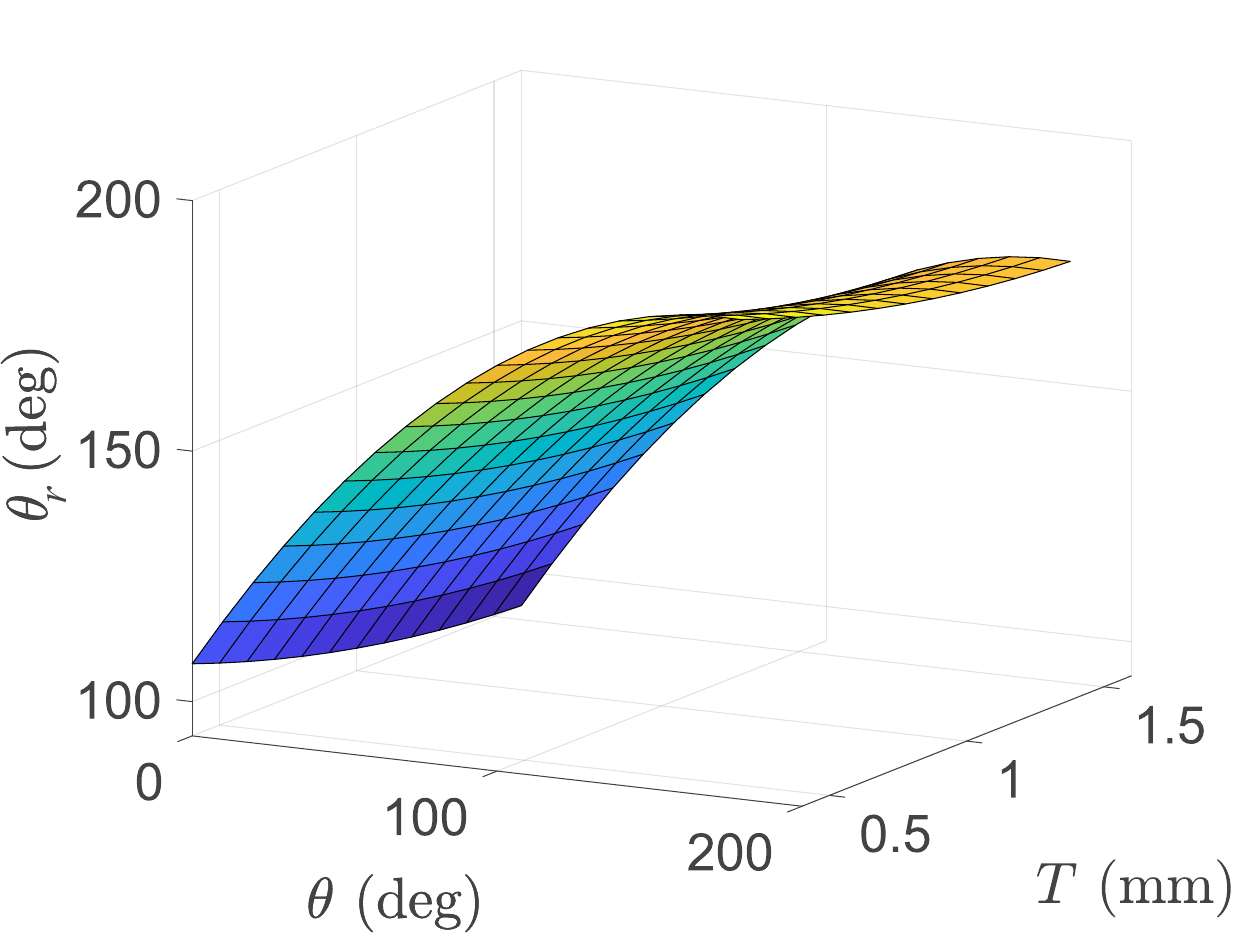} }}%
    \caption{The trained GP model for angle $\theta$ and joint thickness $T$ inputs and output of body force $F_b$ and return angle $\theta_r$ of curve wave joint.}%
    \label{fig:gaussiancruveforce}%
\end{figure}
By using Gaussian regression, a much better fit was calculated that allowed for more accurate force prediction when it was required in the design stage. As seen in Fig. \ref{fig:gaussiamwigglereturn1}, the fit follows a much more accurate form to the data, without creating oscillation as seen in the polynomial equations's fitting model. It is also easier to reach the point of self-contact for the wave design at 150\degree where there is a slight force increase compared to the angle. Fig. \ref{fig:gaussiamwigglereturn1} also clearly shows that the joint can be rotated freely until a full 90\degree bend has been achieved at which point the material will begin to yield.

The Gaussian regression model as shown in Fig. \ref{fig:gaussiancruveforce} can effectively encapsulate the recorded data and serve as a model for calculating the expected force from angle deformation and joint thickness. However, the regression model becomes obsolete for angle values above 150\degree or below 30\degree due to a lack of recorded data in these regions, particularly the extremely high force values observed in the $T=$1.2 mm and $T=$1.6 mm tests.

 \begin{figure}[t!]
    \centering
    \includegraphics[width=0.32\linewidth]{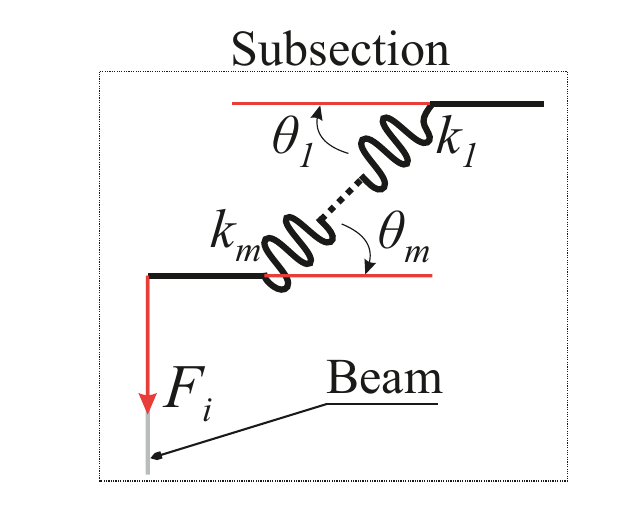}
    \includegraphics[width=0.31\linewidth]{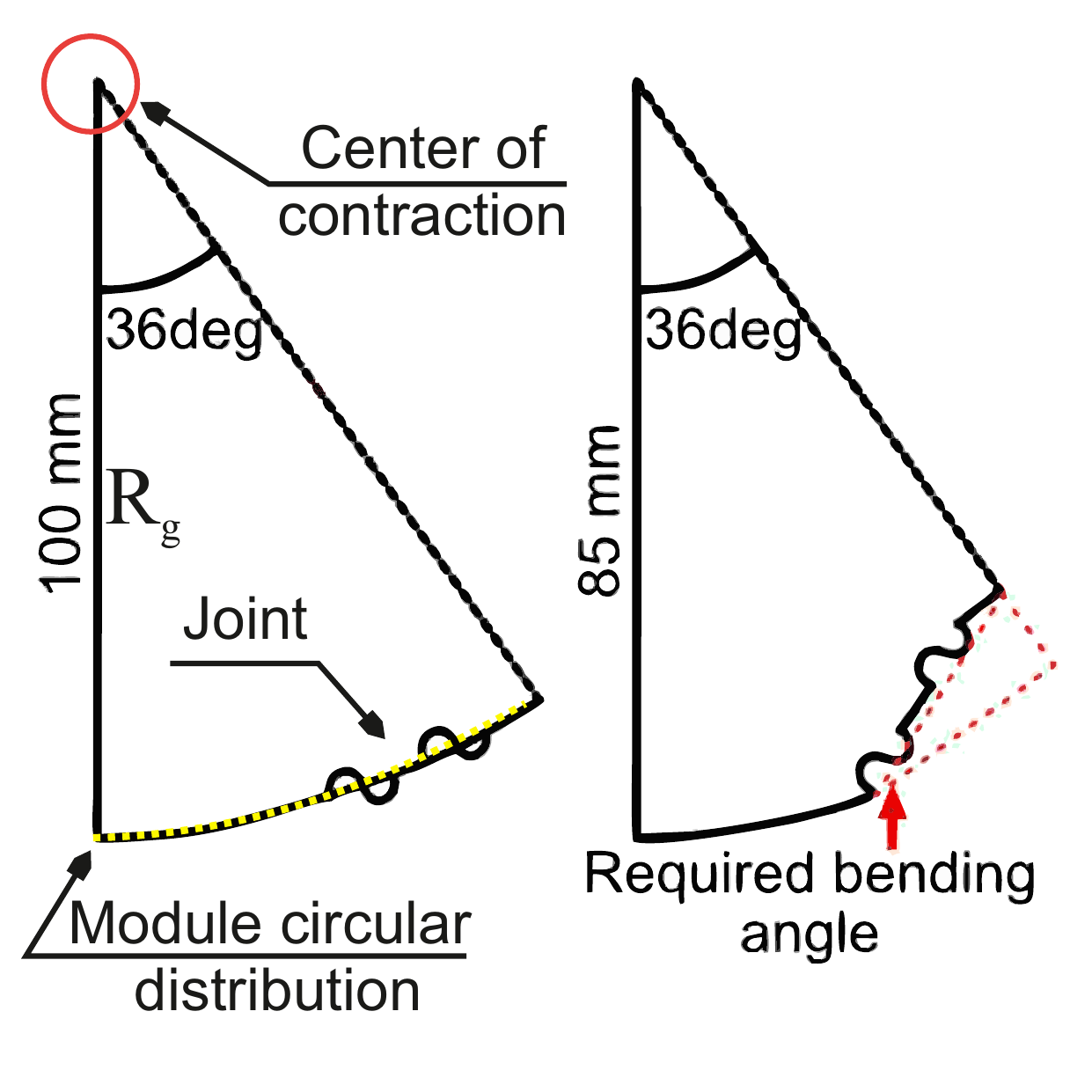}\\
    (a)   \hspace{4.5 cm} (b)
    \caption{(a) The modelled series of joint connections (b) Diagram showing angle required to be bent to achieve a reduction in diameter for expected UGC module.}
    \label{fig:trianglediagram}
\end{figure}
\subsection{Modeling the Circular Modules with Joint Connections}
To obtain a generalised calculation for connected distributed geometric compliant joints, the required torque from motor $\tau_m$ for $n$ sections of the connected beam/cables, we can define it as a summation of left and right subsections force, $F_i=F_l+F_r$ (see example of Fig. \ref{fig:flatprototypeactuated}), which results in
\begin{equation}
    \tau_m=\frac{F_m}{r_m}=\frac{1}{r_m}\left[\sum^n_{i=1}F_i \right],
    \label{Eq:torquemotor}
\end{equation}
where $r_m$ is the radius of the centralized rotational actuator. Please note that spring deformation, both axial and angular, complicates the stiffness behaviour and radius changes. Thus, the force required will differ based on the radius of the approximated circle of module $R_g$. Next, to determine each section's force $F_i$ (having symmetric subsections) using Fig. \ref{fig:trianglediagram}(a), with a series of $i$ number of springs $k_i$, resulting in the following equations
\begin{equation}
 F_i=F_l+F_r\approx2\left[ \sum^{m}_{i=1}\left( \frac{k_i \Delta\theta}{R(t)}\right)\right],
 \label{Eq:forceinsetion}
\end{equation}
where $\Delta \theta=\theta_{i}-\theta_{i-1}$ and $R(t)\in[r_m, \;R_g]$ are the relative angular change between the first, last spring, and changing outer radius of the module. Also, note that each spring stiffness is $k_i=\tau_i/\Delta \theta=(R(t)F_i)/\Delta \theta$.

\section{Underactuated Geometric Compliant Module}
\label{Sec:UGCCompliantmodule}
In this section, we present our finalized single 3D printed UGC modules and compare their results. Lastly, we discuss the successful development of our UGC module results and analyze their behaviour.
\subsection{Passive Design and Tests}
\begin{figure}[t!]%
    \centering
    \includegraphics[width=7cm]{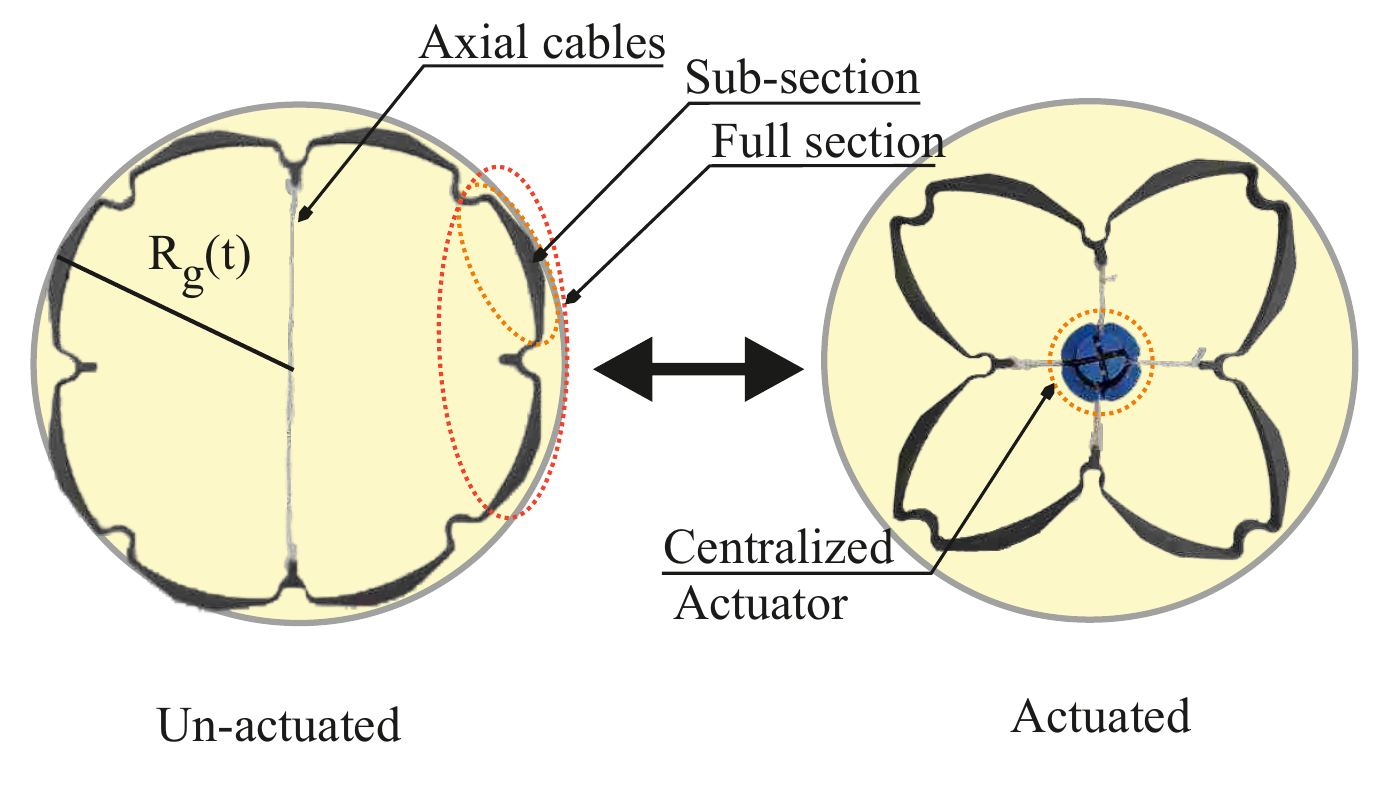} 
    \caption{First prototype using curved joints in singular distribution form.}%
    \label{fig:prototypecurvedjoints}%
\end{figure}

Our first design is shown in Fig. \ref{fig:prototypecurvedjoints}, used curved joints with two cable connections in the centre. This design required a large amount of force to actuate and did not decrease in radius efficiently. This is due to the shape change that caused part of the geometry to bloom outwards while the rest moved in. Our targeted aim is to reduce the $R_g$ outer radius of the module based on a contraction of beams/cables. 

\begin{figure}[t!]%
    \centering
    \subfloat[\centering Actuated]{{\includegraphics[trim={12cm 5cm 12cm 0},clip,height=3.2cm]{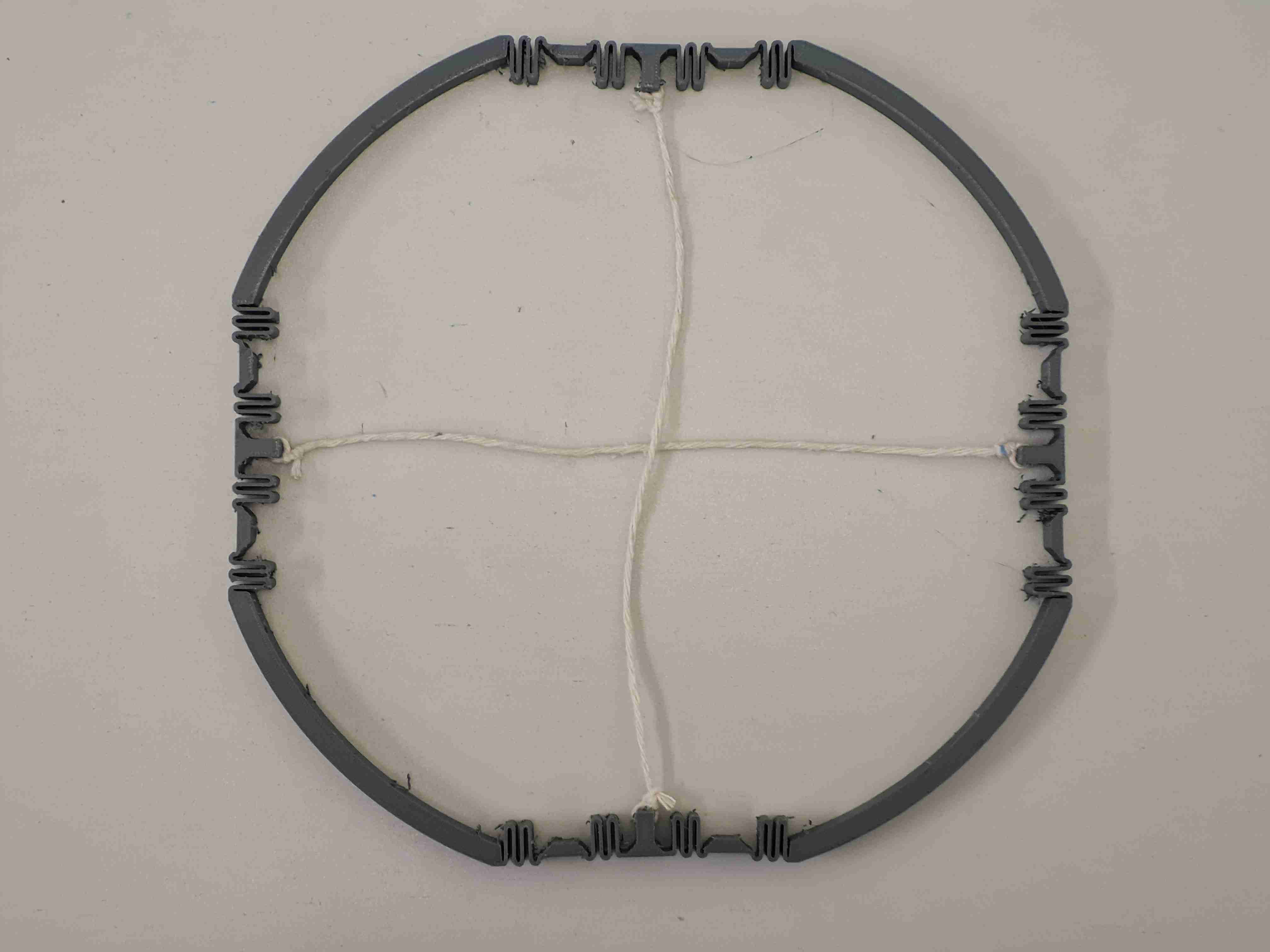} }}%
    \qquad
    \subfloat[\centering Actuated side view]{{\includegraphics[height=3.2cm]{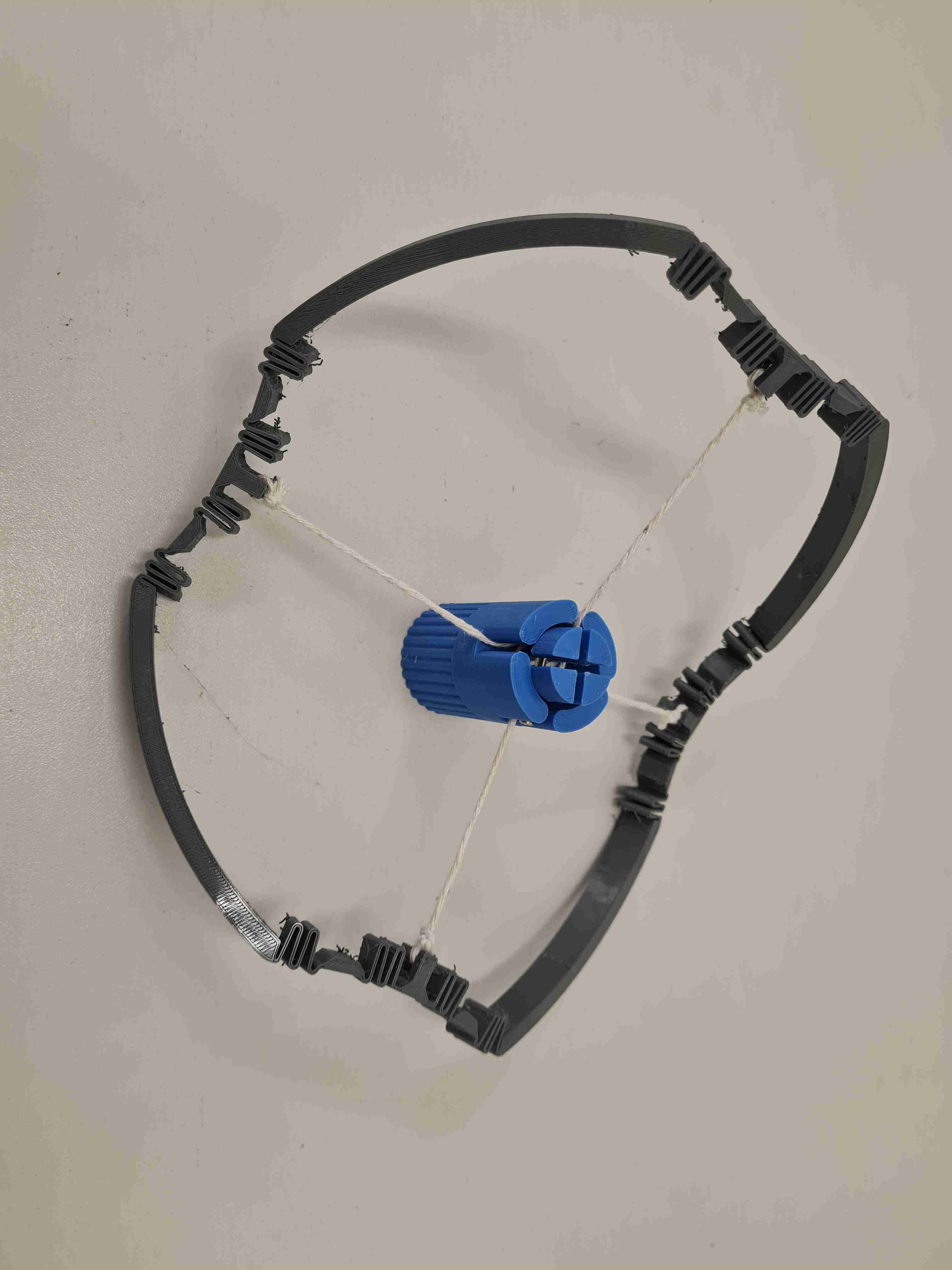} }}%
    \caption{Passive prototype of passive actuation with series of square wave model.}%
    \label{fig:flatprototypeactuated}%
\end{figure}

The next design used a series of square wave joints in linear series to create collapsible areas on the ring as shown in Fig. \ref{fig:flatprototypeactuated}. In this first prototype, by pulling in on the cables, these sections would be pulled into the centre, essentially removing that area of the circumference (compressing the module) and pulling the outer rings closer together. This way, there should be a minimal deformation in the main geometry. This design would also require less force in order to achieve the same reduction in radius as the previous design. However, as seen in Fig. \ref{fig:flatprototypeactuated}, during actuation, due to the lower force requirements, the geometry failed to maintain structural rigidity which caused extreme 3D arching. This would be missing multiple layers of these rings together to constrain any movement in the vertical axis and create non-symmetric compression of the module which is another important observation from the designed UGC module. 



\begin{figure}[t!]%
    \centering
    \subfloat[\centering Un-actuated]{{\includegraphics[width=5cm]{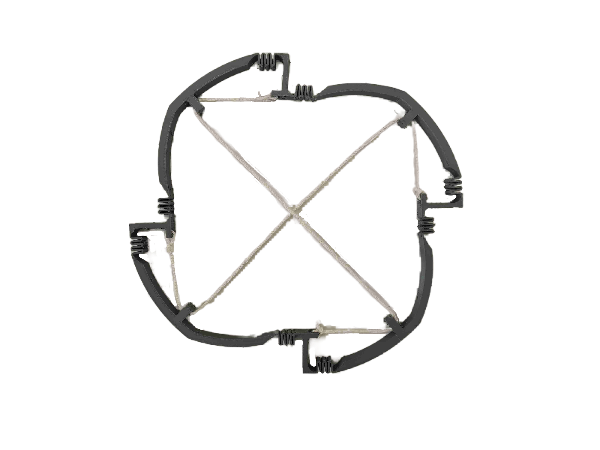} }}%
    \qquad
    \subfloat[\centering Actuated]{{\includegraphics[width=2.5cm]{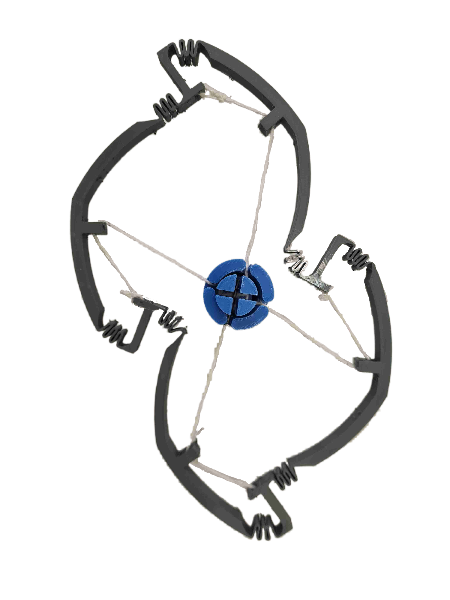} }}%
    \caption{Prototype pulley design with perpendicular square wave models.}%
    \label{fig:pulleyprototype}%
\end{figure}

Another passive compression of UGC design is in Fig. \ref{fig:pulleyprototype}, which was briefly tested using a pulley system in order to drag a section of the out layer under a previous layer. The cable would contact the outer shell and be redirected perpendicular to a connection piece located on a neighbouring section. This design did function but required constant manual rotational from the centralised actuator location to maintain its shape. Such a design is not naturally stable in a circular position when actuated, due to the inconstancy actuation method and high dependency on order of contraction from cables. The redirection also added friction in the system which greatly increased the force required for it to function.

\subsection{Active Design and Evaluation}
\begin{figure}[t!]%
    \centering
    \subfloat[\centering Un-actuated]{{\includegraphics[height=3.4cm]{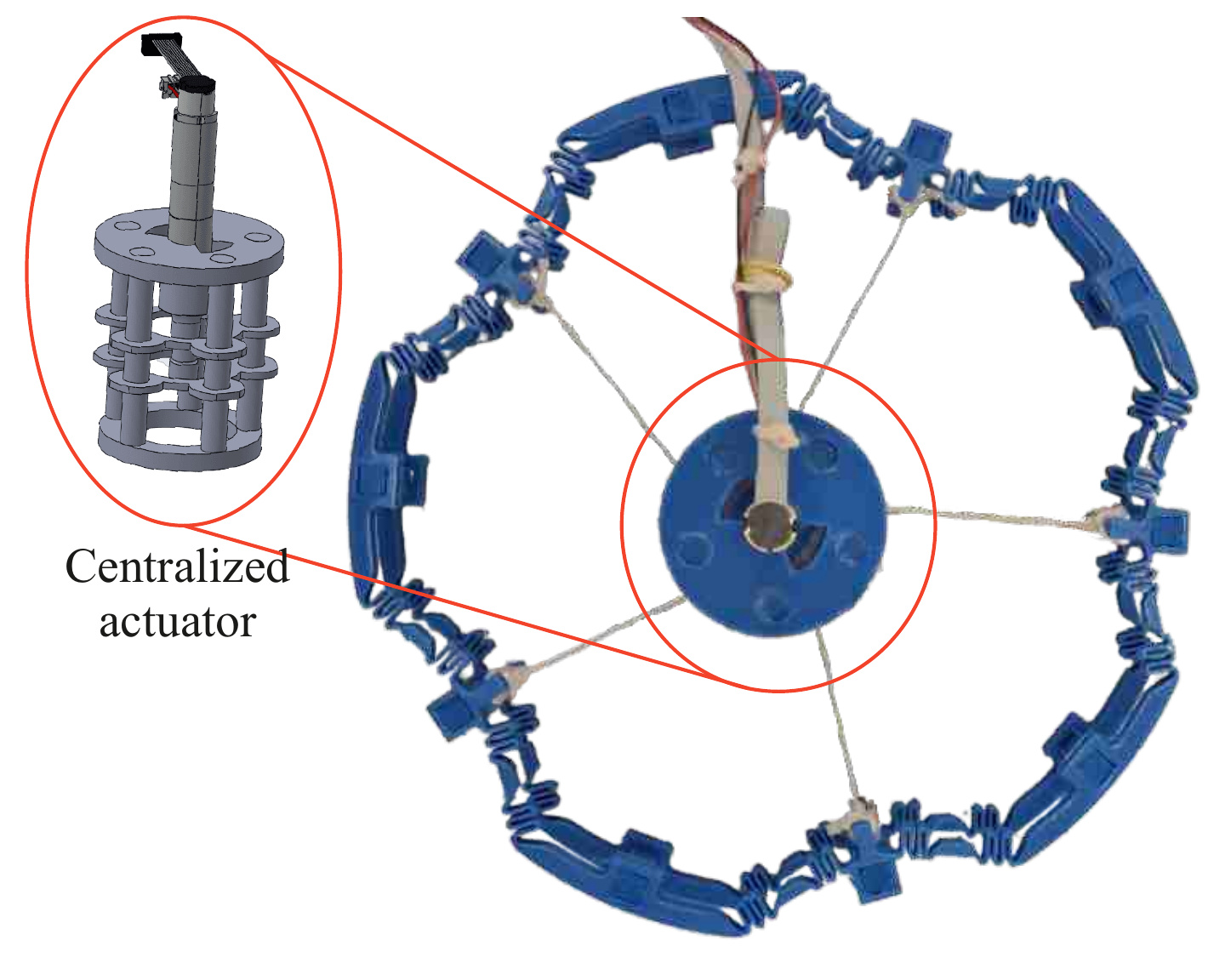} }}%
    \qquad
    \subfloat[\centering Actuated]{{\includegraphics[height=3.1cm]{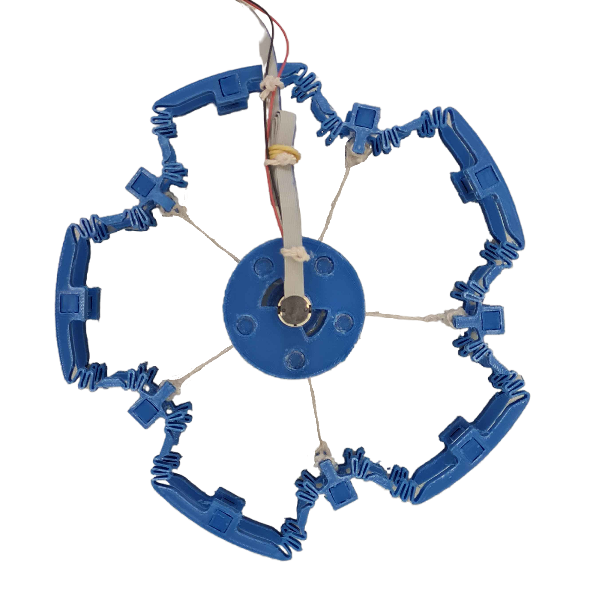} }}%
    \caption{Showing both actuate and not actuated states showing 80-85\% radius change}%
    \label{fig:compressionshownfinalDesign}%
\end{figure}

For the final successful prototype featuring an actuator as the UGC module, the objective was to achieve approximately an 80-85\% reduction in diameter to demonstrate the effectiveness of UGC models. Based on hierarchical test designs and experiments conducted with passive modules, it was determined that employing a 5-section ring (as depicted in Fig. \ref{fig:compressionshownfinalDesign}) would enable more consistent deformation throughout the ring, drawing from our experience with the perpendicular square wave mode (illustrated in Fig. \ref{fig:pulleyprototype}). The initial diameter was set to be 200mm mostly due to the limitation of the print bed of 3D printer. In order to prevent the bending that appeared in previous prototypes, two layers of these rings were printed and connected using vertical connectors that would slot together and keep the centralized actuator properly balanced in centre of the modular total body. This would also allow for internal space inside the structure to better integrate the motor mounting components.

To calculate the force required to actuate the ring design, the number of joints and the angle required for each joint, derived from Eq. (\ref{Eq:torquemotor}) at the motor actuator end, are essential. This total force aids in determining the spindle diameter to prevent motor overload. The ring, depicted in Fig. \ref{fig:trianglediagram}, consists of 5 mirrored repeating modules, allowing analysis of each 36\degree section individually as Fig. \ref{fig:trianglediagram}-b. In its uncompressed state, each section has a radius of 100mm and an arc distance of 62.8mm. To reduce the radius to around 85\%, the arc distance must decrease to 53.4mm, resulting in a reduction in circumference per section. The change in shape can be simplified into a triangle, where the hypotenuse represents half the original arc length, and the adjacent length is the arc distance minus the required change. With these lengths, the required bending angle can be determined, indicating that all joints need to bend accordingly for the module diameter to decrease. Although this triangle simplification is an estimate, it provides insight into the required angle. By knowing the angle requirements, the required motor torque can be determined by calculating (\ref{Eq:torquemotor})-(\ref{Eq:forceinsetion}). At this angle, the return angle is also provided, with negligible damage expected at these forces as expected from GPR model (\ref{Eq:Gaussian_reg}) as shown in Fig. \ref{fig:gaussiamwigglereturn1}.

In the detail of computation, across both rings, there are a total of 40 joints, each requiring 1.05N of actuation force, resulting in a total force requirement of $F_m\approx42N$ from (\ref{Eq:forceinsetion}). The motor utilized was a single 6V DC Maxon motor (making the model underactuated), with a torque of 0.08Nm. Dividing the torque by the force yields a spindle radius of 2mm. To prevent torsional damage to the spindle, the prototype spindle was printed with 100\% infill and a radius of $r_m=3$ mm. This approach was deemed acceptable due to several factors: slightly overshoot on voltage of the motor to enable shorter bursts of increased torque to compensate, and slightly modifying the wave joints with smoother transitions to enhance their performance, thereby facilitating smoother operation. This control problem in future will be carefully studied to have successful feedback control to bring the $R_g$ to specific radiuses $R(t)$. The spindle housing is directly connected to the motor, serving as a cable redirect, ensuring that the cable emanates from as central a point as possible and preventing the housing from freely spinning instead of actuating the module. With the correct spindle radius, the motor successfully actuated the prototype to the required 80-85\%.
\section{Conclusion}
This study first examines different compliant geometric joints. Subsequently, we introduce a novel underactuated geometric compliant (UGC) module capable of adjusting its radius. The final prototype successfully achieved a reduction to 80-85\% of its initial value. Moreover, the research meticulously analyzes various geometries for compliant joints, optimizing the design to minimize actuation components while enhancing strength and actuation consistency. These achievements demonstrate the efficacy of these methods, highlighting their significance in integrating soft and hard robotics within hybrid systems. This integration particularly benefits snake-like robots, enhancing their adaptive capabilities to conform to the surrounding environment.

We plan to integrate the developed UGC into compact forms for the creation of shape and size-changing snake robots. We will also explore the scalability of these joint in different materials e.g., silicon and TPE, designs, both by decreasing and increasing their sizes, for various robotic platforms.

%
%
%
 \bibliographystyle{splncs04}
 \bibliography{reference}

\end{document}